\documentclass[runningheads]{llncs}

\usepackage[mobile,year=2024,ID=6854]{eccv}

\usepackage{eccvabbrv}

\usepackage{graphicx}
\usepackage{booktabs}
\usepackage{xcolor,colortbl}
\usepackage{multirow}
\usepackage{amssymb}
\usepackage{pifont}
\usepackage[accsupp]{axessibility}  

\usepackage{makecell}
\usepackage{svg}
\usepackage{xcolor}
\usepackage{subcaption}
\usepackage{wrapfig}

\usepackage{algorithm}
\usepackage{algpseudocode}
\usepackage{tablefootnote}

\usepackage{mathtools}
\usepackage{textcomp}
\usepackage{amsmath}
\usepackage{bm}

\newcommand{\encoder}{\mathcal{E}}

\newcommand{\bx}{\mathbf{x}}
\newcommand{\bz}{\mathbf{z}}

\usepackage[pagebackref,breaklinks,colorlinks,citecolor=eccvblue]{hyperref}

\usepackage{orcidlink}

\definecolor{yzybest}{rgb}{0.96, 0.57, 0.58}
\definecolor{yzysecond}{rgb}{0.98, 0.78, 0.57}
\definecolor{yzythird}{rgb}{1.0, 1.0, 0.56}
\definecolor{myblue}{HTML}{1f77b4}
\definecolor{myorange2}{HTML}{ff7f03}
\definecolor{myred}{HTML}{FF0100}
\definecolor{gscolor}{rgb}{1.0,0.6,0.0} %

\definecolor{mygray}{gray}{0.9}

\begin{document}

\title{Hybrid Video Diffusion Models with 2D Triplane and 3D Wavelet Representation}
\titlerunning{HVDM}

\author{Kihong Kim \textsuperscript{\rm 1}, 
Haneol Lee \textsuperscript{\rm 2}, 
Jihye Park \textsuperscript{\rm 3},  
Seyeon Kim \textsuperscript{\rm 3},   \newline 
Kwanghee Lee \textsuperscript{\rm 1},
Seungryong Kim $^\dag$\textsuperscript{\rm 3}, and 
Jaejun Yoo $^\dag$\textsuperscript{\rm 2}} 

\authorrunning{Kim et al.}

\institute{\textsuperscript{\rm 1}VIVE STUDIOS \qquad                     \textsuperscript{\rm 2}UNIST \qquad                            \textsuperscript{\rm 3}Korea University \qquad\\
\url{https://hxngiee.github.io/HVDM/}
}

\maketitle

\begingroup
\renewcommand{\thefootnote}{}
\footnotetext{$^\dag$Corresponding author}
\endgroup

\begin{abstract}
Generating high-quality videos that synthesize desired realistic content is a challenging task due to their intricate high dimensionality and complexity. Several recent diffusion-based methods have shown comparable performance by compressing videos to a lower-dimensional latent space, using traditional video autoencoder architecture. However, such method that employ standard frame-wise 2D or 3D convolution fail to fully exploit the spatio-temporal nature of videos. To address this issue, we propose a novel hybrid video diffusion model, called \textbf{HVDM}, which can capture spatio-temporal dependencies more effectively. HVDM is trained by a hybrid video autoencoder which extracts a disentangled representation of the video including: (i) a global context information captured by a 2D projected latent, (ii) a local volume information captured by 3D convolutions with wavelet decomposition, and (iii) a frequency information for improving the video reconstruction. Based on this disentangled representation, our hybrid autoencoder provide a more comprehensive video latent enriching the generated videos with fine structures and details. Experiments on standard video generation benchmarks such as UCF101, SkyTimelapse, and TaiChi demonstrate that the proposed approach achieves state-of-the-art video generation quality, showing a wide range of video applications (e.g., long video generation, image-to-video, and video dynamics control). The source code and pre-trained models will be publicly available once the paper is accepted.
\end{abstract}
\section{Introduction}
\label{sec:intro}

Video generation has received a lot of attention in the computer vision and graphics fields due to its broad applications in making films, creating animations, simulating environments, etc~\cite{blattmann2023align, khachatryan2023text2video, ceylan2023pix2video, dai2023learning, ajay2023compositional}. Nontheless, video generation remains a chellenging task because of the high-dimensionality and complexity inherent in video data. The primary difficulty lies in producing high-quality video frames that are both visually appealing and dynamically coherent.

By leveraging the recent advancements of image-based diffusion models~\cite{ho2021denoising, ho2022imagen}, a variety of video diffusion methods have emerged~\cite{ho2022video, ruan2023mm, mei2023vidm}. These methods show considerable promise in modeling video distributions and synthesizing visually plausible results. However, they often face efficiency challenge due to operating in pixel space. To address this challenge, several recent latent video diffusion models have been developed, training within a compressed latent space to reduce computational demands~\cite{zhou2022magicvideo, he2022latent, esser2023structure, zhou2022magicvideo, an2023latent}. While these latent diffusion models exhibit improved scaling properties with reduced complexity, the potential of novel autoencoder designs remains largely unexplored. There exist conventional approaches, employing standard 2D or 3D CNN-based autoencoders for video generation~\cite{ho2022video,singer2022make,he2022latent,hu2023lamd}. Yet, these CNN-based architectures struggle with effectively managing the overall spatio-temporal structure of videos. While transformers might be considered as a solution, their direct application to video representation learning can lead to significant computational time and memory inefficiency~\cite{kalchbrenner2017video, weissenborn2019scaling}.

Unlike the aforementioned approaches, PVDM~\cite{yu2023video} offers a novel perspective in video generation. PVDM simplifies video complexity by adopting a triplane representation, where video data is factorized into 2D projected latents across different spatio-temporal directions in a latent space. This method has demonstrated improved generation results over previous works such as StyleGAN-based approaches~\cite{skorokhodov2021stylegan} and other diffusion models~\cite{ho2022video}, particularly in effectively representing intricate video data. Despite its progress, they encounter inherent limitations due to the reduction of higher-dimensional video data into lower-dimensional spaces. More specifically, the 2D projected latents struggle to fully recover the 3D structure of video volume, lacking crucial volume information. 

Motivated by these observations, we propose a novel hybrid video diffusion model designed to comprehensively capture the spatio-temporal dependencies of video, called \textbf{HVDM}. Rather than reinventing the wheel, we leverage these known characteristics to enhance effectiveness in video generation tasks. Our video autoencoder combines transformer for 3D-to-2D projections and the 3D CNNs to effectively utilize both global context and local volume information in video. We first start from a well-known fact that transformers excel at capturing long-range dependencies, and we employ them to acquire the global context of the video. To compensate the limited inductive bias of transformers, we leverage 3D CNNs to provide local volume details, drawing from the recognized advantage of 3D CNNs in capturing short-range spatio-temporal information. More specifically, our proposed autoencoder builds the stengths of both architectures by incorporating global 2D projected and local 3D volume latent representations. These two types of latent representations mutually enhance each other through cross-attention module that operates along both the temporal and spatial axes.

\begin{figure*}[t]
    \centering
    \includegraphics[width=1.0\linewidth]{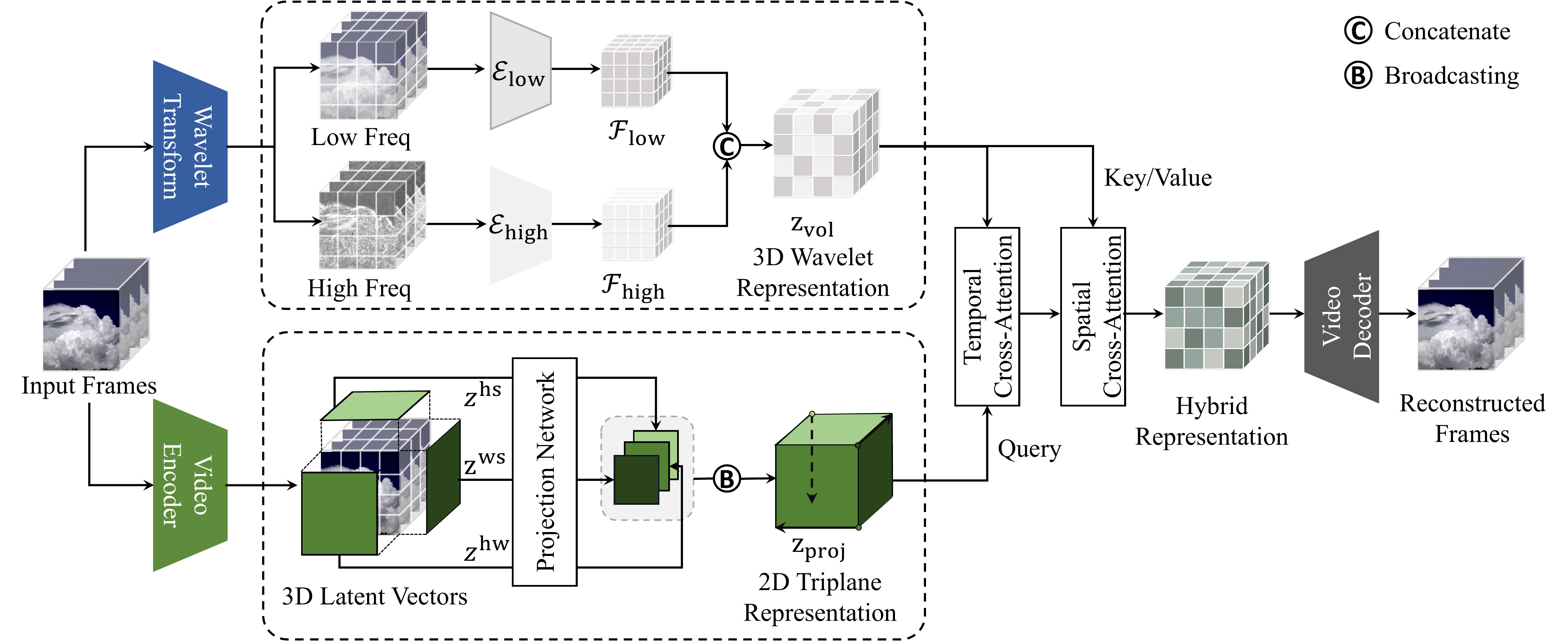}
    \caption{\textbf{Overview of our hybrid video autoencoder in HVDM} that combines a 2D triplane and 3D volume representation for video encoding. The 2D triplane representation provide global context and 3D volume representation provide local volume information of video. The spatio-temporal cross-attention module incorporates these distinctive features to organizes fine-grained video representation.} 
    \label{fig:autoencoder}
    \vspace{-1em}
\end{figure*}
Taking this approach further, we pose the question of whether it is possible to exploit frequency information for boosting the video reconstruction process. We find that employing 3D discrete wavelet transform could potentially resolve the problem of limited receptive fields in 3D CNNs. Specifically, considering that the 3D discrete wavelet transform reduces the video's size by half in the frequency domain, it allows us to expand the receptive field without losing information. Additionally, using 3D discrete wavelet transform decompose complex video into different frequency components at multiple levels of details, providing rich video representation. Finally, by training the video autoencoder with both reconstruction and frequency matching loss in the pixel and spectral spaces, frequency matching loss improved visual quality during video reconstruction process. As a result, we observe that the integration of wavelet-based features provides benefit compared to the use of raw video data by enabling more nuanced video encoding.

To validate the effectiveness of our method, we demonstrate a wide range of video applications including unconditional video generation, long video generation, image-to-video, and video dynamics control. Here, we use three benchmark datasets for video generation: UCF-101~\cite{soomro2012ucf101}, SkyTimelapse~\cite{xiong2018learning}, and TaiChi~\cite{siarohin2019first}. Both quantitative and qualitative results demonstrate video generation capabilities of our method, showing superior performance over the state-of-the-art approaches.
\section{Related Work}
\label{sec:formatting}

\subsubsection{Video generation.}
Video generation methods have been developed based on deep generative models, such as variational autoencoder (VAEs), generative adversarial networks (GANs), and diffusion models (DMs). Early work~\cite{pan2017create, li2018video} attempted to employ VAEs and GANs-based models to generate frames in an autoregressive manner. However, they fail to produce desired results due to the difficulty of modeling spatial-temporal changes and their results limited to low resolutions. Another work~\cite{vondrick2016generating} introduce a spatio-temporal convolutional architecture that effectively disentangles the scene's foreground and background. MoCoGAN~\cite{tulyakov2018mocogan} propose the motion and content decomposed generative adversarial network that generates a video by translating a sequence of random vectors into a sequence of video frames. Although each of these methods show encouraging results for video generation, GAN-based models often suffer from unstable training and fall into mode collapse. In contrast, diffusion models have demonstrated new possibilities for video generation, leveraging a stable training scheme and robust representation.

\subsubsection{Diffusion models.}
Inspired by the success of image-based diffusion models, a series of video diffusion models~\cite{wu2022tune, liu2023video, qi2023fatezero, zhou2022magicvideo, an2023latent, esser2023structure} are proposed. These models leverage the pre-trained image diffusion models and extend their application to videos. While they are capable of efficient learning with just a few computation powers, there are some difficulties in creating consistent and photo-realistic video content. Since they generate videos on a frame-by-frame basis, they face challenges in achieving temporal consistency and producing long video results. 

On the other hand, recent videos diffusion models ~\cite{ho2204video, ruan2023mm, voleti2022mcvd} proposed methods for learning video from scratch by adopting various model architectures. While they are capable of generating high fidelity results that are both temporally consistent and realistic, these methods are still limited in computation and memory inefficiency. To overcome these inefficiency, latent diffusion methods~\cite{he2022latent, yu2023video, hu2023lamd} have emerged which train the model in low-dimensional latent space while maintaining their perceptual quality.

\subsubsection{Triplane representations.}
The effectiveness of triplane representation has already found in 3D-aware generation works~\cite{Schwarz2020NEURIPS, chan2021pi, chan2022efficient, anciukevivcius2023renderdiffusion, wang2023rodin, gu2023nerfdiff}. These triplane representation methods have been shown to offer an effective compression benefits by projecting the 3D volume as 2D-shaped latent features. By taking advantage of these triplane representation, PVDM ~\cite{yu2023video} achieves comparable results in the video generation task, encoding the cubic structure of video pixels. Inspired by this, we present an improved video encoding method that exploit the 2D projected latent with wavelet-based 3D volume latent.
\section{Preliminaries}

\subsection{2D Triplane Representation for Video Diffusion Model}
\label{sec:pvdm}
Given a video sample $\bx \in \mathbb{R}^{3 \times S \times H \times W}$ with time $S$, height $H$, and width $W$, the triplane autoencoder encodes it as three 2D projected latent $[\bz_{\mathrm{proj}}^\mathrm{hw}, \bz_{\mathrm{proj}}^\mathrm{ws}, \bz_{\mathrm{proj}}^\mathrm{hs}]$ where $\bz_{\mathrm{proj}}^\mathrm{hw} \in \mathbb{R}^{C \times H' \times W'}$, $\bz_{\mathrm{proj}}^\mathrm{ws} \in \mathbb{R}^{C \times S \times W'}$, $\bz_{\mathrm{proj}}^\mathrm{hs} \in \mathbb{R}^{C \times S \times H'}$. We denote $C$ as number of channels, $H' = H/d$, $W' = W/d$ for downsampling factors $d > 1$. Unlike conventional video autoencoder which depends on frame-wise 2D or 3D convolution networks~\cite{ho2022video,singer2022make,he2022latent,hu2023lamd}, the PVDM~\cite{yu2023video} effectively decompose the 3D data using video transformer~\cite{bertasius2021space}  for projections $[\bz_{\mathrm{proj}}^\mathrm{hw}$, $\bz_{\mathrm{proj}}^\mathrm{ws}$, $\bz_{\mathrm{proj}}^\mathrm{hs}]$.
The utilization of these latent vectors $[\bz_{\mathrm{proj}}^\mathrm{hw}, \bz_{\mathrm{proj}}^\mathrm{ws}, \bz_{\mathrm{proj}}^\mathrm{hs}]$ offers several advantages in terms of achieving both computation and memory efficiency. Furthermore, their image-like structure enables the efficient video generation from the latent space reducing the dimensionality of the data. 

\subsection{3D Wavelet Representation}
3D Wavelet transform is a classical algorithm widely used in medical imaging and video compression to separate different frequency components within the data~\cite{wang1996medical,wang2007video,secker2002highly}. Just as the 2D wavelet transform, given $\bx \in \mathbb{R}^{3 \times S \times H \times W}$, the 3D wavelet transform operates along $s$, $h$, and $w$ direction decomposing the 3D volume into different subbands. These subbands are factorized into low and high subbands where the low subband express the slowly changing part of the data and the high subband contain fine details capturing sharp edges and rapid transitions within the data.

\begin{wrapfigure}{r}{0.50\linewidth}
    \vspace{-25pt}
    \centering
    \includegraphics[width=1.0\linewidth]{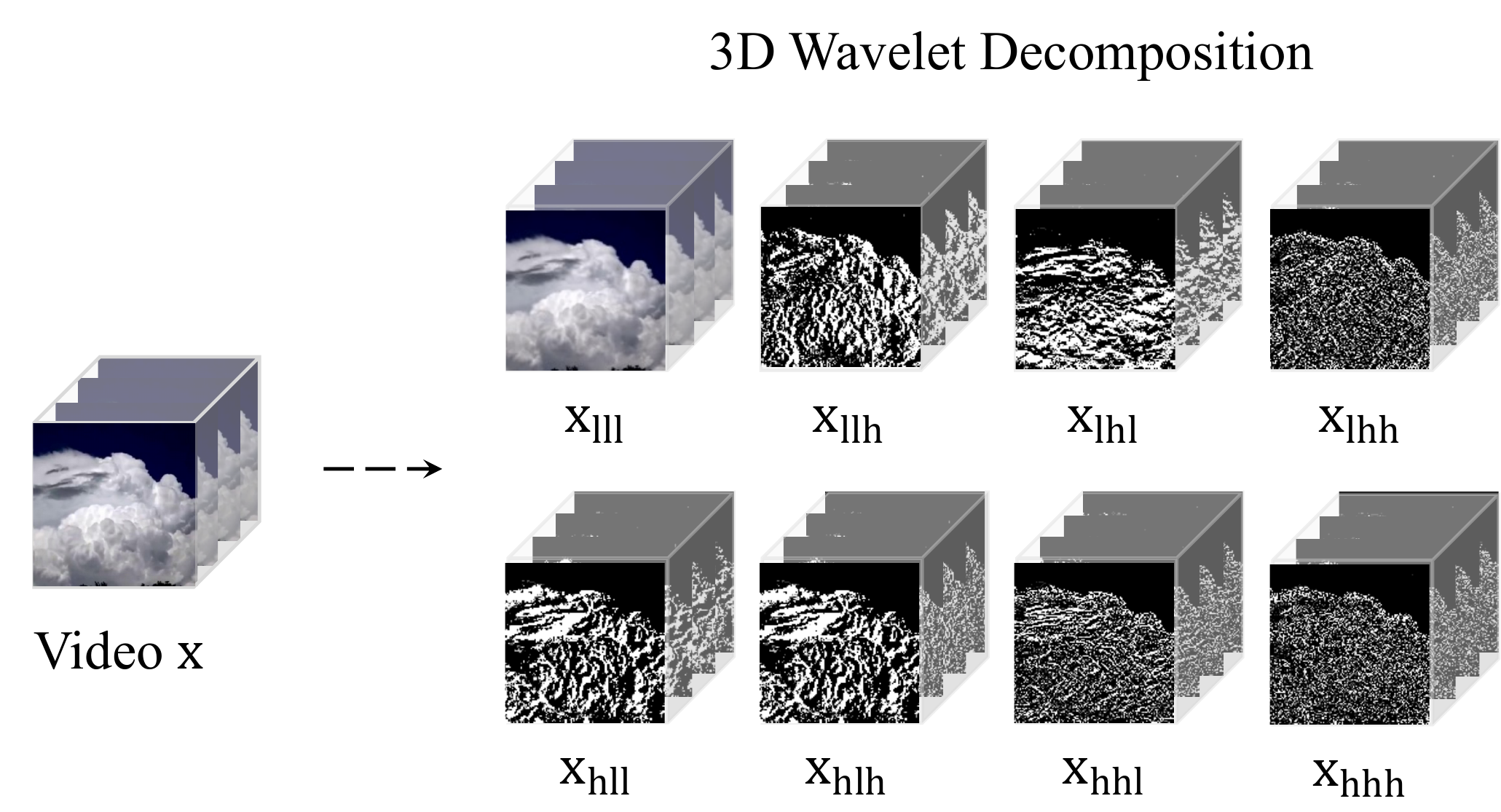}
    \caption{\textbf{Visualization of 3D wavelet transform.} The volume of video is decomposed into eight subband ($\bx_\mathrm{lll}, \ldots, \bx_\mathrm{hhh}$) including low and high frequency components.}
    \label{fig:wavelet_decomposition}
    \vspace{-2em}
\end{wrapfigure}
Specifically, it can be defined as discrete wavelet transform (DWT) and discrete inverse wavelet transform (IWT). In general, Haar wavelets is commonly employed in a variety of real-world applications due to their simplicity. Different types of filters, including low-pass and high-pass filters, divide the volume input $\bx \in \mathbb{R}^{3 \times S \times H \times W}$ into distinct frequency sub-bands $\bx_\mathrm{lll}$, $\bx_\mathrm{llh}$, $\bx_\mathrm{lhl}$, $\bx_\mathrm{lhh}$, $\bx_\mathrm{hll}$, $\bx_\mathrm{hlh}$, $\bx_\mathrm{hhl}$, $\bx_\mathrm{hhh}$ with a size of $S/2 \times H/2\times W/2$. Where the low-frequency components provide the long-term averages information, while the high-frequency components provide information about local changes. In Fig.~\ref{fig:wavelet_decomposition}, we visualize a one level of 3D discrete wavelet decomposition.

\section{Methodology}

Our hybrid video autoencoder is illustrated in Fig.~\ref{fig:autoencoder}. We first describe the overall video encoding architecture in Sec~\ref{sec:autoencoder}. In the following, we present the training objective which enforce constraints in both pixel space and the wavelet space in Sec~\ref{sec:lossfunction}. Then, we formalize the diffusion-based video generation framework in Sec~\ref{sec:DM}. Lastly, we introduce various video applications using video diffusion model in Sec~\ref{sec:video_generator}. 

\subsection{Designing Hybrid Video Autoencoder}
\label{sec:autoencoder}
To train a video diffusion model generating high quality results, it is necessary to have an expressive video representation that accounts for inclusive information of video data. To satisfy this requirement, we construct hybrid representations by combining 2D triplane and 3D wavelet representation. 

\subsubsection{2D triplane representation.}
The 2D projected latent which is originated from PVDM~\cite{yu2023video} is extracted by 3D-to-2D projections mapping at each spatio-temporal direction to encode 3D video pixels. As shown in Sec~\ref{sec:pvdm}, the 2D image-like latent $[\bz_\mathrm{proj}^{\tt hw}, \bz_\mathrm{proj}^{\tt ws}, \bz_\mathrm{proj}^{\tt hs}]$ can be encoded with a combination of a video encoder~\cite{bertasius2021space} and a small transformer~\cite{vaswani2017attention}. Formally, the projected latent is formulated as follows:
\begin{equation}
    \begin{split}
    [u_\mathrm{shw}] &:=\, \mathcal{T}_{\mathrm{video}}^\mathrm{shw}(\bx),
      \\   
    \bz_{\mathrm{proj}}^\mathrm{hw} &:=\, \mathcal{T}_\mathrm{proj}^\mathrm{s}(u_\mathrm{1hw},\ldots,u_\mathrm{Shw}),
    \\
    \bz_\mathrm{proj}^\mathrm{ws} &:=\, \mathcal{T}_\mathrm{proj}^\mathrm{h}(u_\mathrm{s1w},\ldots,u_\mathrm{sH'w}),
    \\
    \bz_\mathrm{proj}^\mathrm{hs} &:=\, \mathcal{T}_\mathrm{proj}^\mathrm{w}(u_\mathrm{sh1},\ldots,u_\mathrm{shW'}),
    \end{split}
\end{equation}

\noindent where $\mathcal{T}_{\tt video}^\mathrm{shw}$ denotes a video-to-3D-latent encoder such as video transformer~\cite{bertasius2021space} and $\mathcal{T}_\mathrm{proj} = [\mathcal{T}_\mathrm{proj}^\mathrm{s}, \mathcal{T}_\mathrm{proj}^\mathrm{h}, \mathcal{T}_\mathrm{proj}^\mathrm{w}]$ denotes a small transformer~\cite{vaswani2017attention} for projections. While projected latent provide a global information of video with a computational efficiency, this transformer architecture projecting a triplane leads to lossy compression due to the lack of local volume information. Moreover, the weak inductive bias of the transformer leads to a slower convergence speed for the model. To overcome these limitations, we transformed the 2D projected latent into a 3D volume and complemented it with the 3D volume representation. Converting a 2D projected latent into a 3D volume $\bz_{\tt proj}$ is formulated as follows:
\begin{equation}
      \begin{split}
      \bz_\mathrm{proj}^\mathrm{s'hw} &:= [\bz_\mathrm{proj}^\mathrm{1hw}, \ldots ,\bz_\mathrm{proj}^\mathrm{Shw}], \\
      \bz_\mathrm{proj}^\mathrm{sh'w} &:= [\bz_\mathrm{proj}^\mathrm{s1w}, \ldots ,\bz_\mathrm{proj}^\mathrm{sH'w}], \\
      \bz_\mathrm{proj}^\mathrm{shw'} &:= [\bz_\mathrm{proj}^\mathrm{sh1}, \ldots ,\bz_\mathrm{proj}^\mathrm{shW'}], \\
      \bz_\mathrm{proj} &:= [\bz_\mathrm{proj}^\mathrm{s'hw} + \bz_\mathrm{proj}^\mathrm{sh'w} + \bz_\mathrm{proj}^\mathrm{shw'}] .
      \end{split}
\end{equation}

\subsubsection{3D wavelet representation.} 3D CNN based volume encoder $\mathcal{E}_\mathrm{low}, \mathcal{E}_\mathrm{high}$ extract local spatio-temporal features in videos. These volumetric local features are exploited with the 2D projected latent $\bz_\mathrm{proj}$ to reorganize fine-grained features. Instead of utilizing the unprocessed RGB video frame directly in the volume encoder, we employ the decomposed wavelet features as input. These decomposed wavelet features extracted from the 3D wavelet transform $\mathcal{W}$ provide coarse and detailed volume coefficient suited for video encoding. Extracting 3D wavelet latent can be formulated as follows:
\begin{equation}
    \begin{split}
    \mathcal{F}_\mathrm{low} &= \mathcal{E}_\mathrm{low}(\bx_\mathrm{lll}),   \\
    \mathcal{F}_\mathrm{high} &= \mathcal{E}_\mathrm{high}({\text{Cat}}(\bx_\mathrm{llh}, \cdots, \bx_\mathrm{hhh})),  \\
    \bz_\mathrm{vol} &= {\text{Cat}}(\mathcal{F}_\mathrm{low},\mathcal{F}_\mathrm{high}).
    \end{split}
    \label{eq:freq_feature}
\end{equation}
\noindent where $\bx_\mathrm{lll}$ denotes decomposed low-pass subband and the others $\bx_\mathrm{llh}, \ldots, X_\mathrm{hhh}$ denotes high-pass subband. The low-pass subband and high-pass subband are encoded to provide the coarse content and fine motion information.

\subsubsection{Fusing 2D triplane and 3D wavelet representations.}
As shown in Fig.~\ref{fig:autoencoder}, these 2D projected and 3D volume latent are integrated by cross-attention operation across the temporal and spatial axis. In this way, we can reinforce the volume-related inductive bias and rearrange fine-grained video latent for video generation. Formally, this cross-attention process can be formulated as follows:
\begin{equation}
    \begin{split}
        \bz^{i} &= 
        {\text{CA}}(\bz_\mathrm{proj}^{i}, \bz_\mathrm{vol}) + \bz_\mathrm{proj}^{i}, \\
        \bz^{i} &= {\text{FFN}}(\bz^{i}) + {\bz}^{i}, \\
        {\tilde{\bz}^{i}} &= 
        {\text{CA}}({\bz}^{i}, \bz_\mathrm{vol}) + {\bz}^{i}, \\
        \bz^{i+1} &= {\text{FFN}}(\tilde{\bz}^{i}) + \tilde{\bz}^{i},
    \end{split}
\end{equation}
where ${\text{CA}}(a,b)$ refers to Cross-Attention, with a representing the query and b representing the key and value. Additionally, FFN stands for FeedForward Network. Note that the input $\bz_\mathrm{proj}^{0} = \bz_\mathrm{proj}$ and the final output $\bz$. Then, we reconstruct the video $\tilde{\bx}$ with video decoder $\mathcal{D}$ based on the refined hybrid features.
\begin{equation}
        \tilde{\bx} = \mathcal{D}(\bz),
\end{equation}
where $\bz \in \mathbb{R}^{C\times S \times H' \times W'}$.

\subsection{Loss function}
\label{sec:lossfunction}
Our hybrid video autoencoder consists of three loss functions. In the following, we provide a brief overview and explain the training loss functions. Our training loss functions enable the preservation of frequency information during the video reconstruction process. Specifically, we enforce constraints both in the pixel space and the wavelet space.

\subsubsection{Reconstruction loss.}
The reconstruction loss $\mathcal{L}_{rec}$ denotes the similarity between the original input video and output video reconstructed by a model. The reconstruction loss term enforce the model to learn meaningful features in the encoding phase so that it can accurately reconstruct the input data in the decoding phase. We employ mean absolute error for reconstruction:
\begin{equation}
	\begin{aligned}
		\mathcal{L}_{\mathrm{rec}}=\mathbb{E}_{\mathbf{x} \sim \mathcal{X}}\left[\|\bx-\tilde{\bx}\|_{1}\right].
	\end{aligned}
 \label{eq:Recon}
\end{equation}

\subsubsection{Perceptual loss.}
The perceptual loss measures the difference between the feature map of input and a generated output in terms of perceptual similarity. Typically, this loss entails the extraction of feature maps from a pre-trained neural network. In contrast to reconstruction loss, which focuses solely on the pixel-level differences between videos, perceptual loss takes into account high level features such as textures and structures that humans perceive.
\begin{equation}
	\begin{aligned}
		\mathcal{L}_{\mathrm{lpips}}=\mathbb{E}_{\mathbf{x} \sim \mathcal{X}}\left[\|f(\bx)-f(\tilde{\bx})\|_{1}\right].
    \end{aligned}
 \label{eq:FreqMatchingLoss}
\end{equation}
where $f$ denotes the perceptual feature extractor.

\subsubsection{Frequency matching loss.}
Conventional methods typically rely on pixel-level losses in the spatial-temporal space. Since these loss functions closely match the original videos in pixel domain, the generated video's quality may not be sufficient potentially failing to capture the frequency information. Our frequency matching loss is designed to facilitate the model can learn frequency features with less optimization difficulty. This loss not only helps the model to recover missing frequency components but also enhances the overall reconstruction quality. We define the frequency matching loss as a reconstruction term in the wavelet domain:
\begin{equation}
	\begin{aligned}
		\mathcal{L}_{\mathrm{freq}}=\mathbb{E}_{\mathbf{x} \sim \mathcal{X}}\left
  [\|\mathcal{W}_{\mathbf{d}}(\mathbf{x)-\mathcal{W}_\mathbf{d}(\mathbf{\tilde{x}}})\|_{1}\right], 
  \mathbf{d}\in\{\mathrm{s}, \mathrm{h}, \mathrm{w} \}.
	\end{aligned}
 \label{eq:FreqMatchingLoss}
\end{equation}
\noindent where $\mathcal{W}_\mathbf{d}$ denotes three-dimensional discrete wavelet transform and $\mathbf{d}$ denotes the dimensions from which the wavelet transform is performed.

\subsubsection{Overall loss.}
Taking into account all the losses, the overall loss is formulated as follows:

\begin{equation}
\begin{aligned}
\mathcal{L} =  \lambda_{\mathrm{rec}}\mathcal{L}_{\mathrm{rec}}  + \lambda_{\mathrm{lpips}}\mathcal{L}_{\mathrm{lpips}} + \lambda_{\mathrm{freq}}\mathcal{L}_{\mathrm{freq}}, 
\end{aligned}
\end{equation}
 where $\lambda_{\mathrm{rec}} = \lambda_{\mathrm{lpips}} = 1$ and $\lambda_{\mathrm{freq}} = 0.5$ in our experiments. 

Note that, unlike previous works~\cite{yu2023video,blattmann2023align} that focus on improving the temporal dynamics of generated videos through the use of a 3D discriminator, our video autoencoder achieves temporal realism in the reconstructed videos without the need for adversarial training.

\subsection{Diffusion Model-based Video Generator}
\label{sec:DM}
Leveraging the hybrid latent extracted from the video autoencoder, we train a diffusion-based video generator which learn the latent distribution $p(\mathbf{z})$ through the reverse process of the Markov Chain.

Our diffusion model consists of two processes: the forward process and the reverse process. In forward process, the model gradually transforms the latent $\mathbf{z}_0 \sim p(\mathbf{z}_0)$ toward a Gaussian distribution as formulated below:
\begin{equation}
\begin{split}
    q(\bz_{t}|\bz_{t-1}) &= \mathcal{N}(\bz_t; \sqrt{1-\beta_t}\bz_{t-1}, \beta_t\mathbf{I}), \\ q(\bz_t| \bz_0) &= \mathcal{N}(\bz_t; \sqrt{\bar{\alpha}_t}\bz_0, (1-\bar{\alpha}_t)\mathbf{I}).
\end{split}
\end{equation}
where $\bar{\alpha}_t = \prod_{i=1}^{t} (1-\beta_i)$ and $\beta$ represents a pre-defined schedule for the variance. 

In the reverse process, the model is trained to gradually denoise the latent via the parameterized Gaussian transition.
\begin{equation} \label{eq:forward_ddpm}
    p_\theta(\mathbf{z}_{t-1}|\mathbf{z}_t)=\mathcal{N}(\mathbf{z}_{t-1}; \mu_\theta (\mathbf{z}_t, t), \sigma^2_t\mathbf{I}),
\end{equation}
\begin{equation}
    \mu_\theta\left(\mathbf{z}_t, t\right)=\frac{1}{\sqrt{\alpha_t}}\left(\mathbf{z}_t-\frac{\beta_t}{\sqrt{1-\bar{\alpha}_t}} \epsilon_\theta\left(\mathbf{z}_t, t\right)\right),
\end{equation}
\noindent where $\epsilon_\theta(\cdot)$ is a trainable denoising diffusion model which recover a noisy sample $\mathbf{z}_t$. The denoising diffusion model is trained to predict the noise with the following simplified objective:
\begin{equation}
   \mathcal{L}_\mathrm{simple}(\theta):= \left\| \epsilon_\theta(\bz_t, t) - 
   \epsilon \right\|_2^2,
\end{equation}
where $\bz_t = \sqrt{\bar{\alpha}_t}\bz_0 + \sqrt{1-\bar{\alpha}_t}\epsilon$.

To generate video samples from the latent space, we train a denoising diffusion model which follow the standard 3D U-Net architecture. The 3D U-Net architecture is composed of 3D convolution layers combined with skip connections. We employ a simplified 3D U-Net to denoise the latent $\bz$ for the purpose of verifying the effectiveness of our video autoencoder.

\begin{figure*}[t!]
\centering
\includegraphics[width=1.0\textwidth]{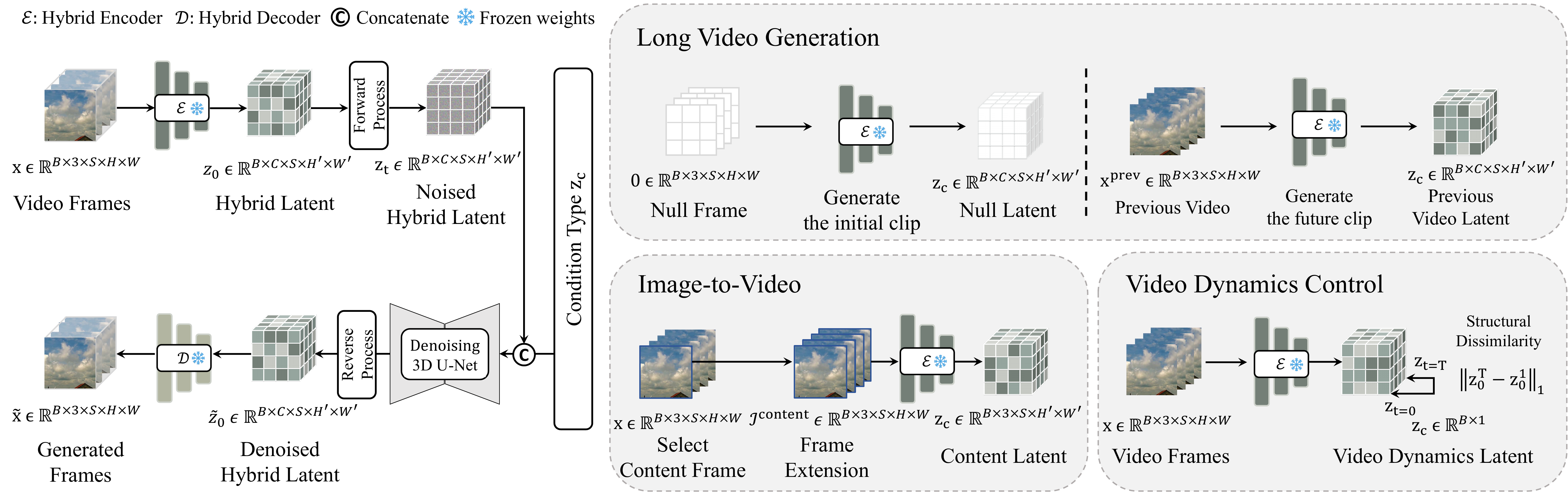}
\caption{
\textbf{Applications for diverse video generator by our proposed HVDM.} Our HVDM is adaptable for diverse video generation tasks depending on the type of condition latent $z_c$. During training process, the condition latent $z_c$ is extracted from the video and jointly trained with the noised hybrid video latent $z_{t}$. During the sampling process, these condition latent are supported to enable various video generation tasks, such as long video generation, image-to-video, and video dynamics control.}
\label{fig:process}
\vspace{-10pt}
\end{figure*}

\subsection{Applications for Diverse Video Generator}
\label{sec:video_generator}
We demonstrate that our method can be extended to a variety of video applications such as long video generation, image-to-video, and video dynamics control. 

\subsubsection{Long video generation.} To facilitate long video generation, our work follows the traditional frame prediction approach~\cite{voleti2022mcvd, he2022latent, yu2023video} where previous frame conditions are combined with future frames. Given two consecutive video clips $[\bx^\mathrm{prev}, \bx^\mathrm{next}]$ of $\bx^\mathrm{prev}, \bx^\mathrm{next} \in \mathbb{R}^{3 \times S \times H \times W}$, we can train the model to generate initial clips and future clips in a unified manner. More specifically, we can generate an initial clip with a null latent and extend the video by composing the condition with previous frame latent as follow:
\begin{equation}
\bz_{c}^\mathrm{long} = \begin{cases}
\encoder(\bx^\mathrm{prev}), &~\text{if future clip}, \\
0, &~\text{otherwise},
\end{cases}
\end{equation}
\noindent where $\encoder$ denotes our pre-trained hybrid encoder.

\subsubsection{Image-to-video.}
Image-to-video generation aims to synthesize a coherent video starting from an image. In general, videos are composed of a series of consecutive images, where the first frame serves as a representative depiction of the overall content within the video. To facilitate the process of image-to-video generation without introducing an additional image encoder, we repeatedly stack the first frame of the video on the temporal axis. Then, we use the latent $\mathcal{I}^\mathrm{content}$ extracted from our video encoder as the content condition. By leveraging the latent $\mathcal{I}^\mathrm{content}$ which capture the common content, we can effectively guide the model to generate coherent video sequence as follow:
\begin{equation}
\begin{split}
\mathcal{I}^\mathrm{content} &= [\bx_0^1, \ldots, \bx_0^1] \in \mathbb{R}^{3 \times S \times H \times W}, \\
\bz_{c}^\mathrm{img2vid} &= \encoder(\mathcal{I}^\mathrm{content}).
\end{split}
\end{equation}

\subsubsection{Video dynamics control.} Empowering users to control motion intensity in video creation provides flexibility with an improved user experience. Unfortunately, it was difficult to obtain motion-related labels from existing video collections for training diffusion models. To enable the alteration of motion intensity, we devise a method that learns motion dynamics with inter-frame variance. Specifically, we define the level of motion alteration as the structural dissimilarity between the semantic latent of the start and end frames. Since the structural dissimilarity encompasses both semantic and motion variance, we employ it as guiding cues to control motion without explicit motion labeled video. Formally, our structural dissimilarity is defined as follow:
\begin{equation}
\bz_{c}^\mathrm{dynamics} = \|\bz_0^{\tt T}-\bz_0^{\tt 1}\|_1.
\end{equation}

Inspired by the classifier-free guidance techniques, we train a single diffusion model to jointly learn both the unconditional distribution $p(\bx)$ and the conditional distribution $p(\bx^{2} | \bx^{1})$. In particular, we train the conditional distribution $p(\bx^{2} | \bx^{1})$ with condition-specific information (e.g., previous frame or content or dynamics latent) and the unconditional distribution using a null condition (e.g., $\bz_c = 0$). The optimized objective can be formulated as follow:
\begin{equation}
\mathbb{E}_{\bx_0, \epsilon, t}
\Big[ \lambda||\epsilon - \epsilon_{\theta}(\bz_t, \bz_c, t)||_2^2 + (1-\lambda)||\epsilon - \epsilon_{\theta}(\bz_t, {0}, t)||_2^2
\Big],
\end{equation}
where $\bz_0\,{=}\, \encoder({\bx_0}), \bz_t \,{=}\, \sqrt{\bar{\alpha}_t}\bz_0 \,{+}\, \sqrt{1-\bar{\alpha}}_t{\epsilon}$ and $\lambda \,{\in}\, (0, 1)$ is a hyperparameter that guide a model to learn an unconditional and conditional distributions.
\section{Experiments}

\subsection{Training Details}
\noindent\textbf{Datasets.}
We conduct the experiments on three benchmark datasets for the video generation: UCF-101~\cite{soomro2012ucf101}, SkyTimelapse~\cite{xiong2018learning}, and TaiChi~\cite{siarohin2019first}. UCF-101~\cite{soomro2012ucf101} contains 13,320 videos from 101 action classes, offering a diverse range of human actions. SkyTimelapse~\cite{xiong2018learning} is composed of 5,000 videos containing dynamic sky scenes. TaiChi~\cite{siarohin2019first} consists of full human bodies performing Tai-Chi actions. 

\begin{table*}[t]
\centering 
\resizebox{\linewidth}{!}
{
\begin{tabular}{l c c c c c c c c c c c c}
\toprule
& \multicolumn{4}{c}{UCF-101}
& \multicolumn{4}{c}{SkyTimelapse}  
& \multicolumn{4}{c}{TaiChi}  
\\
\cmidrule(lr){2-5}  \cmidrule(lr){6-9}  \cmidrule(lr){10-13}
Method 
& {$\text{R-FVD}$  $\downarrow$}
& {$\text{LPIPS}$  $\downarrow$}
& {$\text{PSNR}$  $\uparrow$}
& {$\text{SSIM}$  $\uparrow$}

& {$\text{R-FVD}$  $\downarrow$}
& {$\text{LPIPS}$  $\downarrow$}
& {$\text{PSNR}$  $\uparrow$}
& {$\text{SSIM}$  $\uparrow$}

& {$\text{R-FVD}$  $\downarrow$}
& {$\text{LPIPS}$  $\downarrow$}
& {$\text{PSNR}$  $\uparrow$}
& {$\text{SSIM}$  $\uparrow$}

\\
\midrule
    LVDM ~\cite{he2022latent}
    &{-} &{-} &{-} &{-}
    &{74.92} &{0.091} &{31.21} &{0.778}   
    &{102.68} &{0.112} &{27.63} &{0.729}  \\
    
    PVDM ~\cite{yu2023video}  
    &{27.03} &{0.095} &{27.03} &{0.742}
    &{35.96} &{0.061} &{32.71} &{0.826}
    &{61.61} &{0.068} &{26.87} &{0.715}  \\
    \midrule
    HVDM (ours)   
    &{\textbf{5.35}}  &{\textbf{0.038}} &{\textbf{34.00}} &{\textbf{0.915}}    
    &{\textbf{20.76}} &{\textbf{0.046}} &{\textbf{35.04}} &{\textbf{0.891}}   
    &{\textbf{21.51}} &{\textbf{0.025}} &{\textbf{33.51}} &{\textbf{0.916}}  \\
    \bottomrule
\end{tabular}}
\vspace{5pt}
  \caption{\textbf{Quantitative comparison of reconstruction results.} Our video autoencoder shows significantly superior performance across various metrics, providing more fine-grained comparision.}
  \label{tab:recon_table}
\end{table*}

\begin{figure*}[h!]
\vspace{-15pt}
\centering
\includegraphics[width=0.90\textwidth]{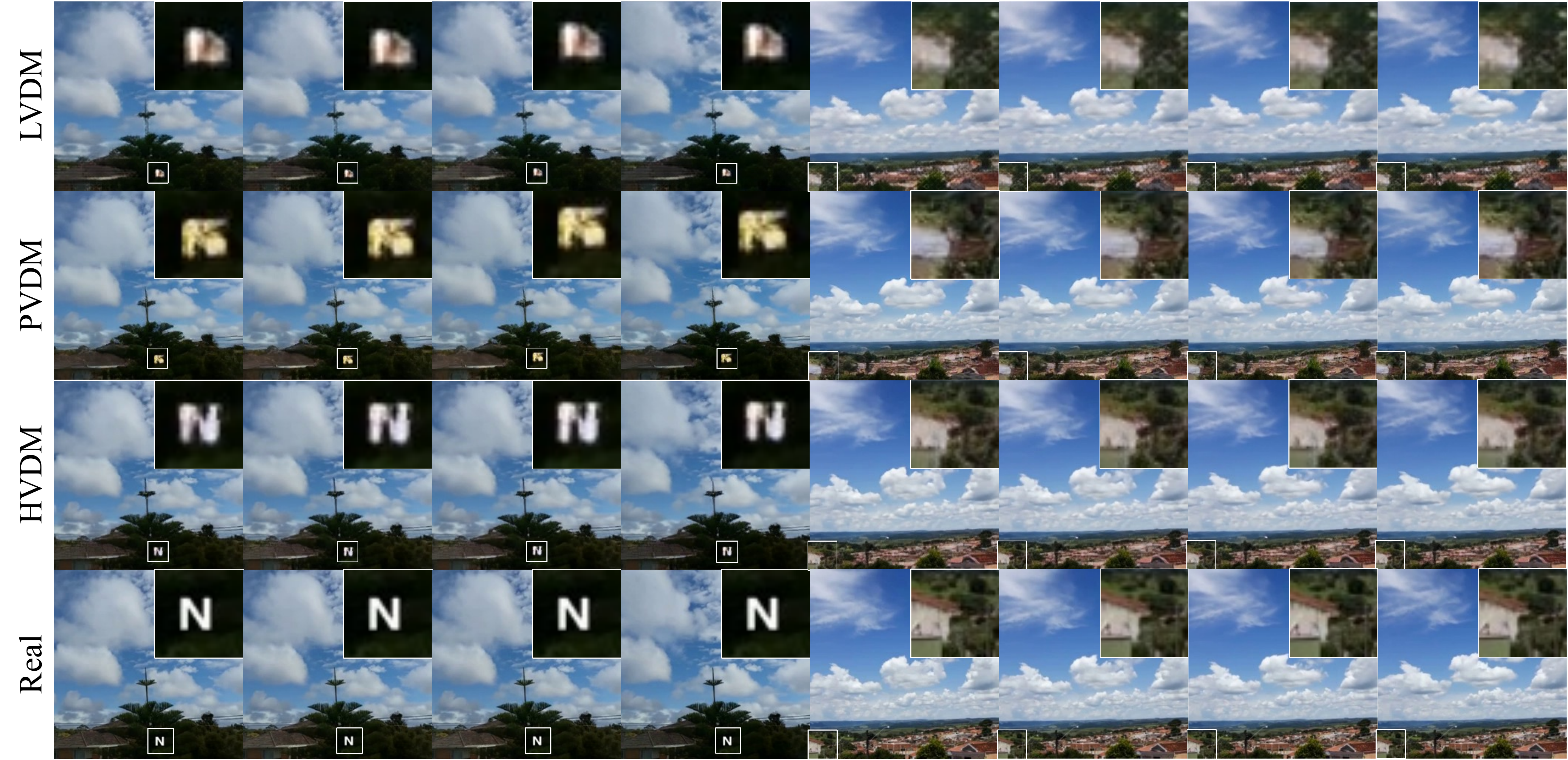}
\caption{\textbf{Qualitative reconstruction results on SkyTimelapse~\cite{xiong2018learning} dataset~\cite{siarohin2019first}.} Our HVDM produces distinct and sharp edge details and faithfully expresses a fine structures in a video sequence. Additional samples are included in appendix.} 
\label{fig:REC}
\vspace{-15pt}
\end{figure*}

\noindent\textbf{Training.}
The video autoencoder component for 2D projected latent stem from the PVDM~\cite{yu2023video}, but the other wavelet-based frequency component is added. We train our models on 4 NVIDIA A100 GPUs for each autoencoding and diffusion process. See Appendix \textcolor{red}{A} for the details.

\subsection{Evaluation Details}
\noindent\textbf{Baseline models.}
We compare HVDM with the following models: VideoGPT~\cite{yan2021videogpt}, MoCoGAN~\cite{tulyakov2018mocogan}, MoCoGAN-HD~\cite{tian2021good}, DIGAN~\cite{yu2022digan}, StyleGAN-V~\cite{skorokhodov2021stylegan}, LVDM~\cite{he2022latent}, and PVDM ~\cite{yu2023video}. We utilize official model weights and conduct training using the official code when the model weights are not available.

\noindent\textbf{Quantitative evaluations.}
To evaluate reconstruction quality, we use R-FVD~\cite{Unterthiner2019FVDAN}, LPIPS~\cite{zhang2018perceptual}, PSNR~\cite{huynh2008scope}, and SSIM~\cite{hore2010image}. Here, R-FVD denotes FVD between reconstructed videos and the ground-truth videos. As for the evaluation of generation quality, we measure FVD~\cite{Unterthiner2019FVDAN} and KVD~\cite{unterthiner2018towards}. Considering different clip lengths, we denote FVD$_{16}$ and KVD$_{16}$ scores, as well as FVD$_{128}$ and KVD$_{128}$ scores for video clips with a length of 16 and 128. For a fair comparison, we follow the evaluation protocol of StyleGAN-V~\cite{skorokhodov2022stylegan}; i.e., we use 2,048 real/fake video clips for evaluating FVD. See Appendix \textcolor{red}{B} for the details.

\begin{table*}[t!]
  \centering
  \resizebox{1\linewidth}{!}{
  \begin{tabular}{lccccccccc}
  \toprule
  & VideoGPT
  & MoCoGAN
  & + StyleGAN2
  & MoCoGAN-HD
  & DIGAN
  & StyleGAN-V
  & LVDM$^\dagger$
  & PVDM$^\dagger$
  & HVDM (ours) \\
  & \multicolumn{9}{c}{\cellcolor{gray! 20} \text{Train Split}} \\
  \midrule
  \text{UCF101} - {$\text{FVD}_{16}$} & 2880.6 & 2886.8 & 1821.4 & 1729.6 & 1630.2 & 1431.0 & N/A & 399.4 & \textbf{303.1} \\
  \text{UCF101} - {$\text{FVD}_{128}$} & N/A & 3679.0 & 2311.3 & 2606.5 & 2293.7 & 1773.4 & N/A & \textbf{505.0} &  549.7 \\
  \text{SkyTimelapse} - {$\text{FVD}_{16}$} & 222.7 & 206.6 & 85.88 & 164.1 & 83.11 & 79.52 & 318.9 & 63.5 & \textbf{42.8} \\
  \text{SkyTimelapse} - {$\text{FVD}_{128}$} & N/A & 575.9 & 272.8 & 878.1 & 196.7 & 197.0 
  & N/A & 134.7 & \textbf{124.9} \\ 
  \bottomrule
  \end{tabular}}

  \vspace{+.5\baselineskip}

  \begin{subtable}{.44\textwidth}
  \raggedleft
  \resizebox{1\linewidth}{!}{\begin{tabular}[t]{lccc}
  \toprule
  & LVDM$^\dagger$ & PVDM$^\dagger$ & HVDM (ours) \\ 
  & \multicolumn{3}{c}{\cellcolor{gray! 20} \text{Train Split}} \\
  \midrule
  \text{UCF101} - {$\text{KVD}_{16}$}        & N/A  & 35.5  & \textbf{23.6} \\
  \text{UCF101} - {$\text{KVD}_{128}$}       & N/A  & 37.7  & \textbf{34.4} \\ 
  \text{SkyTimelapse} - {$\text{KVD}_{16}$}  & 21.73 & 3.1  & \textbf{1.5}  \\
  \text{SkyTimelapse} - {$\text{KVD}_{128}$} & N/A  & 6.0   & \textbf{3.5}  \\
  \bottomrule
  \end{tabular}
  }
  \end{subtable}
  \begin{subtable}{.38\textwidth}
    \raggedright
    \resizebox{1.025\linewidth}{!}{\begin{tabular}[t]{lccc}
    \toprule
    & LVDM$^\dagger$ & PVDM$^*$ & HVDM (ours) \\ 
  & \multicolumn{3}{c}{\cellcolor{gray! 20} \text{Train+Test Split}} \\
    \midrule
    \text{TaiChi} - {$\text{FVD}_{16}$}  & 120.9 & 267.0 & \textbf{77.0}  \\
    \text{TaiChi} - {$\text{FVD}_{128}$} & N/A   & 339.2 & \textbf{258.5} \\
    \text{TaiChi} - {$\text{KVD}_{16}$}  & 14.7 & 66.5   & \textbf{11.9} \\
    \text{TaiChi} - {$\text{KVD}_{128}$} & N/A   & 120.1 & \textbf{77.8} \\
    \bottomrule
    \end{tabular}
    }
  \end{subtable}
  \vspace{10pt}
  \caption{\textbf{Quantitative comparison of video generation models on UCF-101, SkyTimelapse and TaiChi datasets.} We evaluate FVD and KVD metrics (the lower values are better) and the best score are marked as bold. The overall score of other baselines are referenced from StyleGAN-V~\cite{skorokhodov2022stylegan}. Methods marked with $^\dagger$ denote the score is computed using official pretrained weights and $^*$ denote the score is reproduced using official codes. We report evaluation details and sampling settings on Appendix \textcolor{red}{B}.} 
  \label{tab:main-results}
\end{table*}

\begin{figure*}[t!]
\vspace{-15pt}
\centering
\includegraphics[width=\textwidth]{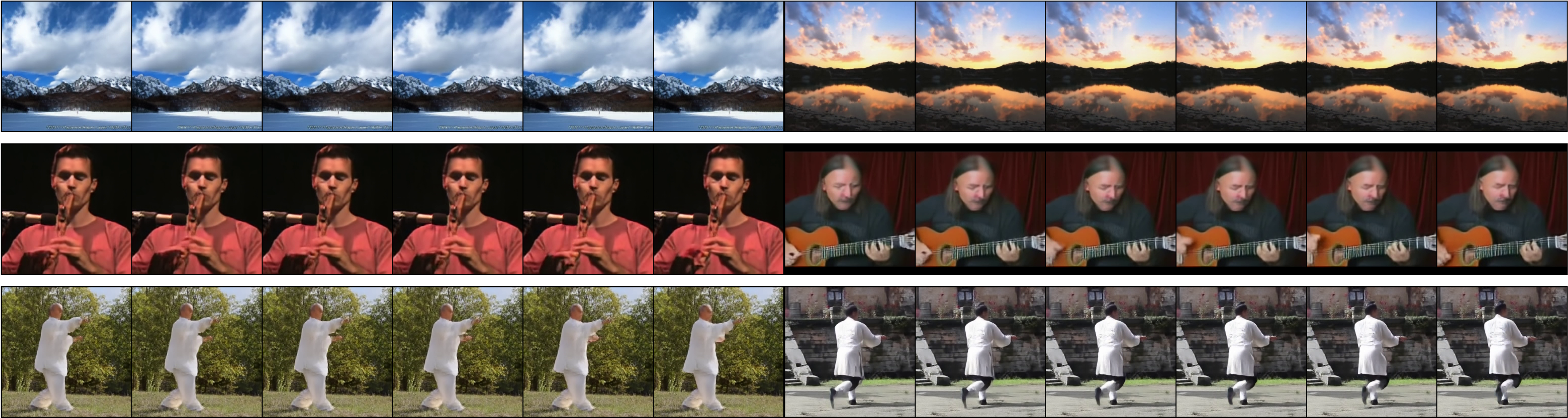}
\caption{\textbf{Generated video sample by our HVDM on UCF-101~\cite{soomro2012ucf101}, SkyTimelapse~\cite{xiong2018learning}, and TaiChi datasets~\cite{siarohin2019first}.}} 
\label{fig:gen}
\vspace{-15pt}
\end{figure*}

\subsection{Comparison Results}
\noindent\textbf{Reconstruction results.} 
Fig.~\ref{fig:REC} and Appendix \textcolor{red}{E.1} show the performance of the reconstructed videos from our autoencoder and other diffusion-based video generative models. Our autoencoder presents a more accurate and clear reconstructed video compared to other baseline methods. In the Tab.~\ref{tab:recon_table}, we demonstrate that our autoencoder achieves superior performance in all metrics for the three datasets.

\noindent\textbf{Generation results.} 
We perform video generation on three datasets and compare the results with the other video generation methods. As shown in Fig.~\ref{fig:gen}, our method is capable of generating not only monotonic motion videos such as SkyTimelapse~\cite{xiong2018learning} and TaiChi~\cite{siarohin2019first} but also dynamic motion video UCF-101~\cite{soomro2012ucf101}, which contain complex action categories. Note that synthesizing high quality videos on a complex dataset like UCF-101~\cite{soomro2012ucf101} has been a challenging problem where existing video generation method often struggle with producing natural motions with good quality. As shown in Tab.~\ref{tab:main-results}, our method achieves superior performance in the benchmark datasets. See Appendix \textcolor{red}{E} for more samples.

\vspace{2em}
\subsection{Interpreting the frequency supervision}
\begin{wrapfigure}{l}{0.50\linewidth}
    \vspace{-20pt}
    \centering
    \includegraphics[width=0.85\linewidth]{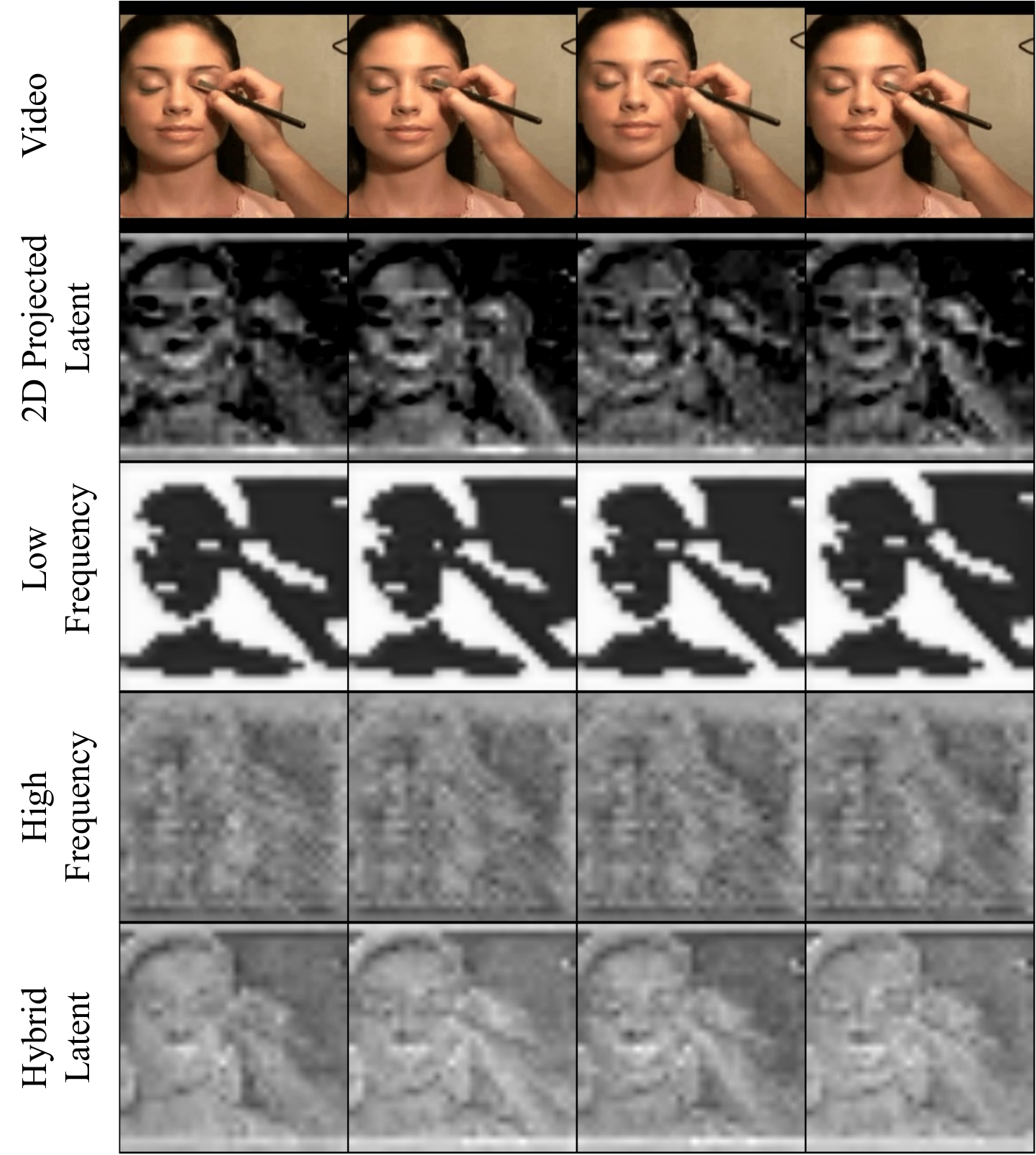}
    \caption{\textbf{Visualization of our hybrid representations.}}
\label{fig:wavelet_decomposition_abl}
    \vspace{-20pt}
\end{wrapfigure}
We perform visualization to understand how hybrid representation can help autoencoder to encode the 3D video pixels in Fig.~\ref{fig:wavelet_decomposition_abl}. The decomposed 3D wavelet features is divided into low frequency and high frequency components, where each component conduct cross-attention operation with the 2D projected latent. As shown in Fig.~\ref{fig:wavelet_decomposition_abl}, low frequency component captures static and dynamic regions and high frequency component capture the sharp edges and important details of video. To refine the ambiguity of the 2D project latent which contain global and abstract information, we incorporate these latent with frequency components. As a result, our autoencoder figures out where the objects are located and how object’s motion occurs in the video.

\subsection{Ablation study on backbone architectures} We conduct verification of performance based on the autoencoder's backbone architecture on UCF-101 dataset. As shown in Tab.~\ref{tab:ae_backbone}, our hybrid autoencoder demonstrates effectiveness in capturing both perceptual details and temporal coherence in the video. We can observe that 3D CNN-based autoencoder show better performance compared to the 2D triplane-based autoencoder since it has more latent dimensions. Although the 2D triplane-based autoencoder effectively compresses latent dimensions, enhancing temporal coherence necessitates adversarial training with a 3D discriminator. Our hybrid autoencoder integrates the strengths of both architectures, achieving superior performance. 
\begin{table}[t]
\centering
\resizebox{0.95\textwidth}{!}{
\begin{tabular}{l c c c c r c c c c c}
\toprule
Backbone  & Proj. &Adv. & $\mathbf{z}$ shape dimension & R-FVD $\downarrow$  & LPIPS $\downarrow$  &PSNR $\uparrow$ &SSIM $\uparrow$ \\
\midrule
{Video Transformer}~\cite{bertasius2021space} &\checkmark &- &$4 \times (32 \times 32 + 16 \times 32 + 16 \times 32)$  &180.21 &0.092 &28.12 &0.782 \\ 
{Video Transformer}~\cite{bertasius2021space} &\checkmark &\checkmark &$4 \times (32 \times 32 + 16 \times 32 + 16 \times 32)$  &{\phantom{0}27.03} &0.095 &27.03 &0.742 \\ 
{3D CNN}~\cite{karpathy2014large} &- &- &$4 \times 16 \times 32 \times 32$          &{\phantom{0}86.73} &0.057 &31.76 &0.852 \\
{3D CNN}~\cite{karpathy2014large} &- &\checkmark &$4 \times 16 \times 32 \times 32$ &{\phantom{0}21.69} &0.064 &30.39 &0.837 \\
\midrule
\textbf{Ours} &\textbf{\checkmark} &\textbf{-} &$\bm{4 \times 16 \times 32 \times 32}$ &\textbf{\phantom{0}\phantom{0}5.35} &\textbf{0.038}  &\textbf{34.00} &\textbf{0.915} \\ 
\bottomrule
\end{tabular}
}
\vspace{10pt}
\caption{\textbf{Ablation study on various backbone architectures}}
\vspace{-20pt}
\label{tab:ae_backbone}
\end{table}

\subsection{Ablation study on fusing modules}
We verify several experiments to assess how the fusing module affects the performance of our hybrid autoencoder on UCF-101 dataset. For feature fusion, we adopted two different modules: 1) the concatenation module 2) the cross-attention module. We also validated the effectiveness of wavelet decomposition by comparing the results which use raw RGB video as input. As shown in Tab.~\ref{tab:ae_fusemodule}, the experimental results show that the cross-attention module and wavelet-decomposed components contribute to the improvement of reconstruction performance. We analyze that utilizing the temporal and spatial attention module facilitate the integration of different 2D triplane and 3D wavelet representation through the comparison of features at different spatial locations.

\begin{figure*}[h]
\vspace{-10pt}
\begin{minipage}[hbt!]{.5\textwidth}
\centering
\resizebox{1.0\textwidth}{!}{
\begin{tabular}{l c c c c c r c c c}
\toprule
Fusion module & R-FVD $\downarrow$ & LPIPS $\downarrow$ &PSNR $\uparrow$ &SSIM $\uparrow$ 
\\
\midrule
Baseline                                & {27.03} & 0.095 & 27.03 & 0.742 \\
Baseline + RGB (Concat.)                & {20.37} & {0.063} & {31.02} & {0.831} \\
Baseline + Low Freq. (Concat.)          & {34.39} & {0.078} & {30.50} & {0.816} \\
Baseline + High Freq. (Concat.)         & {14.96} & {0.061} & {31.32} & {0.838} \\
Baseline + Low \& High Freq. (Concat.)  & {\phantom{0}9.92} & 0.051  &32.69 &0.871 \\
\bottomrule
\end{tabular}}
\label{tab:ae_fusemodule}
\end{minipage}
\begin{minipage}[hbt!]{.5\textwidth}    
\centering
\resizebox{1.0\textwidth}{!}{
\begin{tabular}{l c c c c c r c c c}
\toprule
Fusion module & R-FVD $\downarrow$ & LPIPS $\downarrow$ &PSNR $\uparrow$ &SSIM $\uparrow$ 
\\
\midrule
Baseline + RGB (Cross-attn.)            & {\phantom{0}6.32} & 0.038  &33.79 &0.908 \\   
Baseline + Low Freq. (Cross-attn.)      & {\phantom{0}6.87} & 0.040  &33.69 &0.909 \\   
Baseline + High Freq. (Cross-attn.)     & {\phantom{0}6.97} & 0.047  &33.20 &0.899  \\ 
\midrule
\textbf{Ours} & \textbf{\phantom{0}5.35} &\textbf{0.037}  &\textbf{34.00}  &\textbf{0.915}  \\ 
\bottomrule
\end{tabular}}
\end{minipage}
\captionof{table}{\textbf{Ablation study on feature fusing module.}}
\label{tab:ae_fusemodule}
\end{figure*}

\vspace{-30pt}
\subsection{Ablation study on frequency matching loss}
We validate the effectiveness of frequency matching loss (FML) on UCF-101 dataset in Tab.~\ref{tab:abl_FML}. The result shows that utilizing the FML term leads to a remarkable reduction in the FVD score by approximately 2.73 points demonstrating its effectiveness in enhancing the temporal coherency. Furthermore, evaluation metrics related to perceptual details also improved to 0.003 points for LPIPS, 0.33 points for PSNR, and 0.023 points for SSIM respectively. Based on these results, the FML verifies that aligning frequency components allow the HVDM to enhance temporal changes and perceptual quality of generated videos.

\vspace{-5pt}
\begin{table}[!h]
\centering\small
\resizebox{0.55\textwidth}{!}{
\begin{tabular}{l c c c c c r c c c}
\toprule
Method & R-FVD $\downarrow$ & LPIPS $\downarrow$ &PSNR $\uparrow$ &SSIM $\uparrow$
\\
\midrule
HVDM (w/o FML)             & {8.08}        & {0.041}        & {33.67} & {0.892} \\
\textbf{HVDM (w/\phantom{0}  FML)}  & \textbf{5.35} & \textbf{0.038} & \textbf{34.00} & \textbf{0.915} \\
\midrule
\end{tabular}}
\vspace{5pt}
\captionof{table}{\textbf{Ablation study on frequency matching loss.}}
\label{tab:abl_FML}
\end{table}

\vspace{-3em}
\section{Conclusion}
We proposed a novel hybrid video autoencoder for video generation using diffusion model, dubbed HVDM. We investigated the effectiveness of our hybrid architecture that combines 2D projected and 3D volume representation, leveraging the strengths of frequency matching loss. By incorporating these representations with spatio-temporal cross-attention, HVDM not only address the challenges inherent in video generation, such as the high dimensionality and complexity of video data but also generates high-quality videos with improved realism. Our extensive experiments on benchmark datasets demonstrated the superior performance, compared to the other baseline methods. Furthermore, our approach allows for a wide range of video applications, such as long video generation, image-to-video, and video dynamics control, showing the flexibility and robustness of our method.

%
%
\bibliographystyle{splncs04}
\bibliography{main}

\newpage
\appendix


\clearpage
\appendix

\onecolumn
\begin{center}
{\bf {\Large Hybrid Video Diffusion Models with 2D Triplane and 3D Wavelet Representation}} 
\end{center}
\begin{center}
{\bf {\Large - Supplementary Material -}} 
\end{center}

\section{Implementation Details}
\label{appen:implementation}
In this section we describe details on the model configurations and hyper-parameters used for training of our HVDM. Specifically, for 2D triplane and 3D wavelet representation of our hybrid autoencoder, we modify existing PVDM~\cite{yu2023video} autoencoder architecture: 1) we use reduced depth of each layer in the transformer, 2) we use the decomposed wavelet input instead of the raw video, 3) we use 3D convolution networks to encode the decomposed wavelet features, and 4) we add cross-attention module to fuse 2D projected and 3D wavelet latent. As for the diffusion model, we use a simple 3D convolutional U-Net architecture~\cite{cciccek20163d}. We provide the detailed information for our hybrid autoencoder in Tab.~\ref{appen:ae_config} and U-Net in Tab.~\ref{appen:dm_config}.

\vspace{-1em}

\begin{table}[]
\small
\begin{center}
\resizebox{0.7\textwidth}{!}{
\begin{tabular}{lccc}
\toprule
& UCF-101~\cite{soomro2012ucf101} & SkyTimelapse~\cite{xiong2018learning} & TaiChi~\cite{siarohin2019first}  \\
\midrule
resolution 
& $3  \times 16 \times 256 \times 256$
& $3  \times 16 \times 256 \times 256$
& $3  \times 16 \times 256 \times 256$ \\
base-channels & 384 & 384 & 384 \\
num-res-blocks & 2 & 2 & 2 \\
heads & 8 & 8 & 8 \\
depth & 5 & 5 & 5 \\
$\mathcal{E}_\mathrm{low}$-channels & 768 & 768 & 768 \\
$\mathcal{E}_\mathrm{high}$-channels & 768 & 768 & 768    \\
$z$-shape & $4 \times 16 \times 32 \times 32$ & $4 \times 16 \times 32 \times 32$ & $4 \times 16 \times 32 \times 32$ \\
batch size & 6 & 6 & 6     \\
model size & 63M & 63M & 63M     \\
optimizer & Adam & Adam & Adam      \\
learning rate& $\text{1e-4}$ & $\text{1e-4}$ & $\text{1e-4}$ \\
training iterations & 250k & 200k & 200k \\
\bottomrule
\end{tabular}}
\end{center}
\caption{\textbf{Autoencoder configuration and training details.}}
\label{appen:ae_config}
\end{table}

\vspace{-4em}

\begin{table}[]
\small
\begin{center}
\resizebox{0.7\textwidth}{!}{
\begin{tabular}{lccc}
\toprule
& UCF-101~\cite{soomro2012ucf101} & SkyTimelapse~\cite{xiong2018learning} & TaiChi~\cite{siarohin2019first}  \\
\midrule
resolution 
& $4  \times 16 \times 32 \times 32$
& $4  \times 16 \times 32 \times 32$
& $4  \times 16 \times 32 \times 32$ \\
base-channels & 192 & 192 & 192 \\
num-res-blocks & 2 & 2 & 2 \\
channel-multipliers & [1,2,4,4] & [1,2,4,4] & [1,2,4,4] \\
linear start & 0.0015 & 0.0015 & 0.0015 \\
linear end & 0.0195 & 0.0195 & 0.0195 \\
timesteps & 1000 & 1000 & 1000 \\
batch size & 120 & 120 & 120     \\
model size & 673M & 673M & 673M     \\
optimizer & AdamW & AdamW & AdamW      \\
learning rate& $\text{1e-4}$ & $\text{1e-4}$ & $\text{1e-4}$ \\
training iterations & 400k & 300k & 350k \\
\bottomrule
\end{tabular}}
\end{center}
\caption{\textbf{Our denoising 3D U-Net configuration and training details.}}
\label{appen:dm_config}
\end{table}

\clearpage

\section{Experiment Setup} 
\subsection{Datasets}
To provide a comprehensive evaluation with baseline, we follow the settings of PVDM for UCF-101~\cite{soomro2012ucf101} and SkyTimelapse~\cite{xiong2018learning} datasets and as well as the settings of LVDM~\cite{siarohin2019first} for the TaiChi~\cite{siarohin2019first} dataset. For training the model, we crop and resize all video clips to $256 \times 256$ scale. All video clips are extracted as consecutive 16 or 128 frame lengths from a randomly selected video.

\vspace{1em}

\noindent\textbf{UCF-101} is an action recognition dataset comprising realistic action videos, including 101 action categories. Following the common practice for evaluating unconditional video generation on UCF-101~\cite{soomro2012ucf101}, we use the train split (9,357 videos) for training and test split (3,963 videos) for evaluation.

\vspace{1em}

\noindent\textbf{SkyTimelapse} is a time-lapse dataset containing videos of sky scenes. We conduct data preprocessing by referring to the official code~\cite{xiong2018learning}. Following the evaluation setup employed in StyleGAN-V~\cite{skorokhodov2022stylegan} and PVDM~\cite{yu2023video}, we use the train split for both model training and evaluation.

\vspace{1em}

\noindent\textbf{TaiChi} is a Tai-Chi action dataset containing 3,078 videos in total. For a fair comparison with the LVDM~\cite{he2022latent}, we use the train+test splits for both model training and evaluation. Moreover, we reproduced PVDM~\cite{yu2023video} under the same setup as the model weight was unavailable.

\subsection{Metrics}
For fair comparison, we follows the evaluation protocol of StyleGAN-V~\cite{skorokhodov2021stylegan}. Unlike previous evaluation protocol that preprocess video datasets with fixed-length clips, StyleGAN-V~\cite{skorokhodov2021stylegan} propose to sample video data first and then randomly select fixed-length clips, particularly avoiding bias issues with extremely long videos. As for evaluation metrics, we measure the video generation quality with the following two metrics: Fréchet Video Distance (FVD)~\cite{Unterthiner2019FVDAN} and Kernel Video Distance (KVD)~\cite{unterthiner2018towards}. In addition, we report the reconstruction quality of the autoencoder with the following four metrics: Learned Perceptual Image Patch Similarity (LPIPS)~\cite{zhang2018perceptual}, Peak Signal-to-Noise Ratio~\cite{huynh2008scope} (PSNR), SSIM (Structural Similarity Index)~\cite{hore2010image}, and R-FVD which denote FVD between real and reconstructed samples.

\subsection{Inference settings for evaluations}
We assessed the performance of video autoencoders by evaluating the reconstruction quality, using identical samples from the SkyTimelapse~\cite{xiong2018learning}, UCF-101~\cite{soomro2012ucf101}, and TaiChi~\cite{siarohin2019first} datasets. To compare the performance of our method with state-of-the-art diffusion-based video generative models, we fixed the number of DDIM~\cite{song2020denoising} steps to a constant value of 100. Additionally we extract 2,048 video clips for computing real statistics and generate 2,048 samples for evaluating fake statistics. 

\clearpage

\vspace{1.5em}
\section{Inference Time and Memory}
Comparing recent video diffusion methods~\cite{he2022latent, yu2023video}, we report inference time and memory efficiency in Tab.~\ref{tab:ae_infer}. For a fair comparision, we generate $256 \times 256$ resolution video on a single NVIDIA 3090 GPU with a fixed batch size one. We also sample from models using 100 steps of DDIM~\cite{song2020denoising} to synthesize video clips of 16 and 128 frames. While PVDM~\cite{yu2023video} exhibited computation and memory efficiency, as shown in the Tab.~\ref{tab:ae_infer}, its overall performance tended to be suboptimal. On the other contrary, our method demonstrated superior performance, showing comparable inference time and memory efficiency to other methods which utilize 3D U-Net architecture in diffusion model.

\begin{figure}[h]
\centering\small
\begin{tabular}{lcc}
\toprule
& \multicolumn{2}{c}{Inference (time/memory)}  \\
\cmidrule(lr){2-3}
Length $\rightarrow$ & 16 & 128 \\
\midrule
LVDM~\cite{he2022latent}               
& 8.85/\phantom{.}9.0  & N/A  \\
PVDM~\cite{yu2023video}
& 9.16/\phantom{.}9.6  & 137.8/10.1  \\
HVDM~(Ours) 
& 9.90/11.6 & 151.4/12.1 \\
\bottomrule
\end{tabular}
\vspace{5pt}
    \captionof{table}{\textbf{Comparison of the inference time (Sec) and memory (GB) of various video diffusion models.} N/A denotes that the values cannot be measured because the model is unavailable.}
    \label{tab:ae_infer}
\end{figure}

\vspace{-1em}
\section{Limitations and Future Directions}
Text-to-video generative models are quickly advancing, offering users exciting opportunities to unleash their creativity in new ways. Although we haven't executed experiments on large-scale video dataset for text-to-video generation tasks due to resource limitations, we are believe that our framework is also effective in this task as well.

Recently, Lumiere~\cite{bar2024lumiere} generates the entire temporal duration of the video at once achieving global temporal consistency. In contrast to existing video models which synthesize sub-sampled set of keyframes followed by temporal super-resolution, they train model to learn globally coherent motion by downsampling the video in multiple space-time scales. As a alternative to compressing long video frames without being limited by memory requirements, employing the multi-level wavelet transform as a pooling operation would be an interesting research direction. Furthermore, although HVDM demonstrates potential for efficient and effective video synthesis, further research avenues exist for its enhancement. For example, exploring the design of diffusion model architectures that are specialized for our hybrid latents or wavelet filter banks that provide a highly efficient signal decomposition in the video domain would be an intriguing further research direction.

\clearpage
\section{Additional Results}
\label{appen:additional_results}

\subsection{Video Reconstruction}
\label{sec:compare-video-recon}

\begin{figure}[]
\centering
\includegraphics[width=1.0\textwidth]{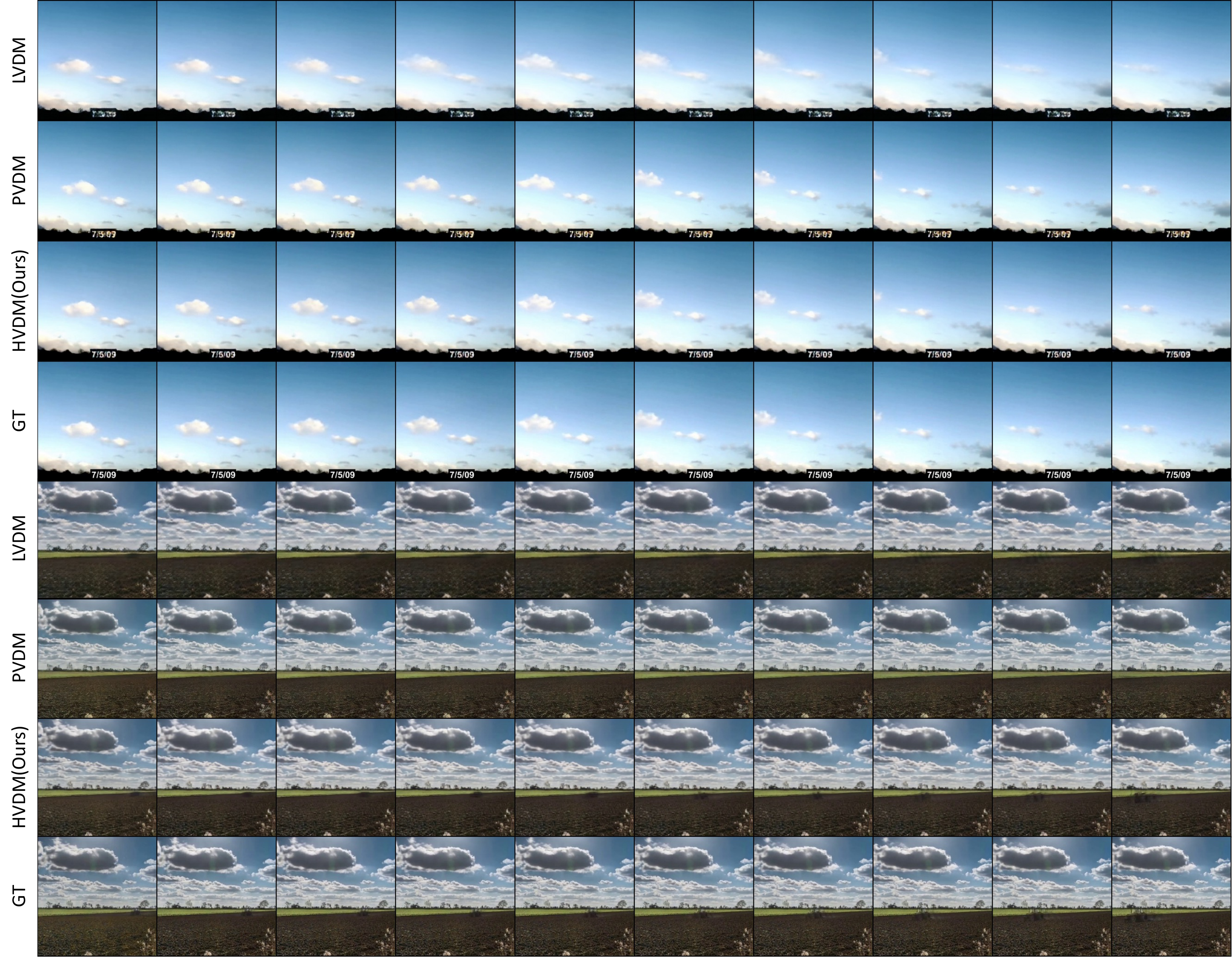}
\captionof{figure}{\textbf{Reconstruction results of our hybrid autoencoder on the SkyTimelapse~\cite{xiong2018learning}.}}
\end{figure}

\clearpage
\begin{figure}[p]
\centering
\includegraphics[width=1.0\textwidth]{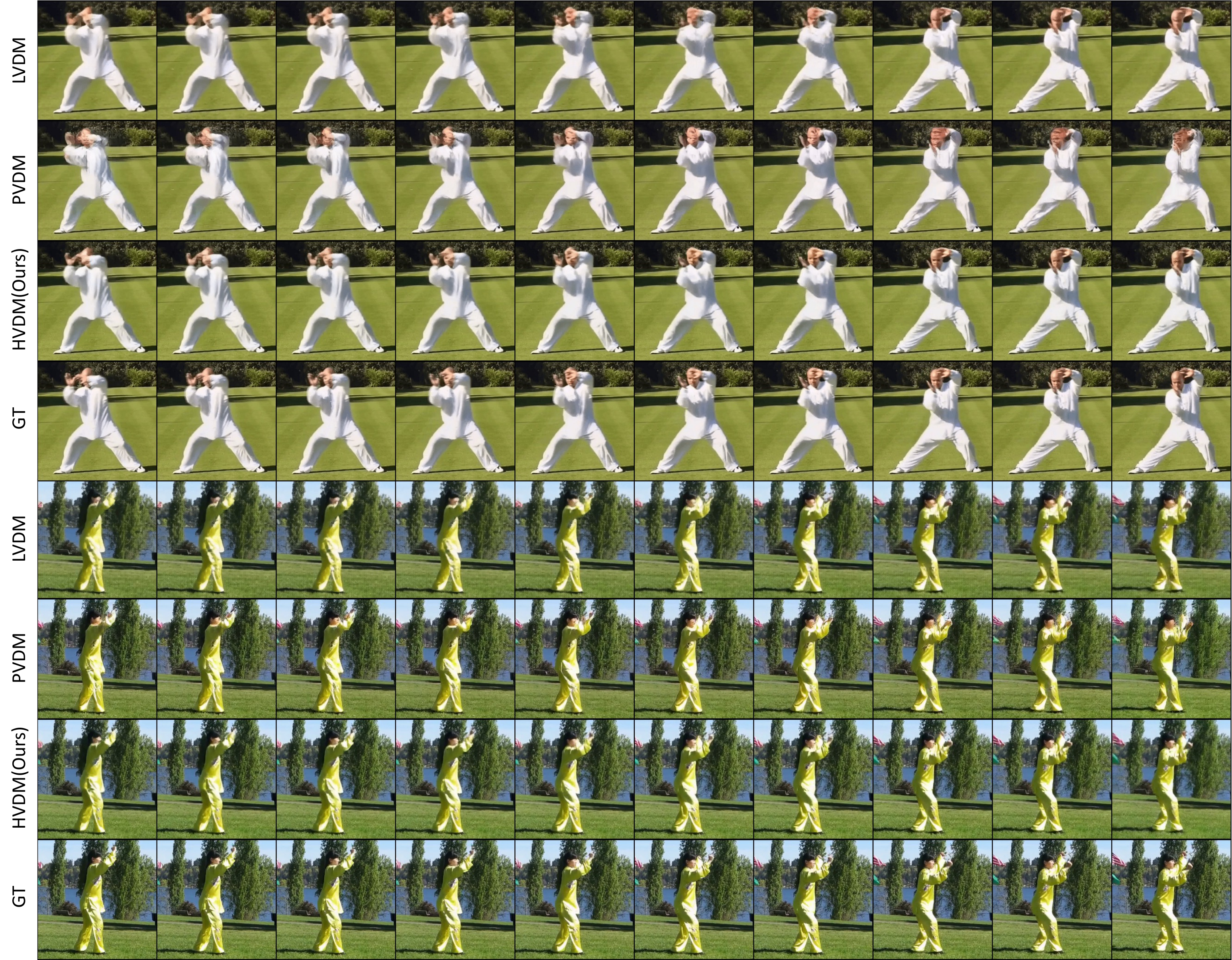}
\captionof{figure}{\textbf{Reconstruction results of our hybrid autoencoder on the TaiChi~\cite{siarohin2019first}.}}
\end{figure}

\clearpage
\begin{figure}[p]
\centering
\includegraphics[width=1.0\textwidth]{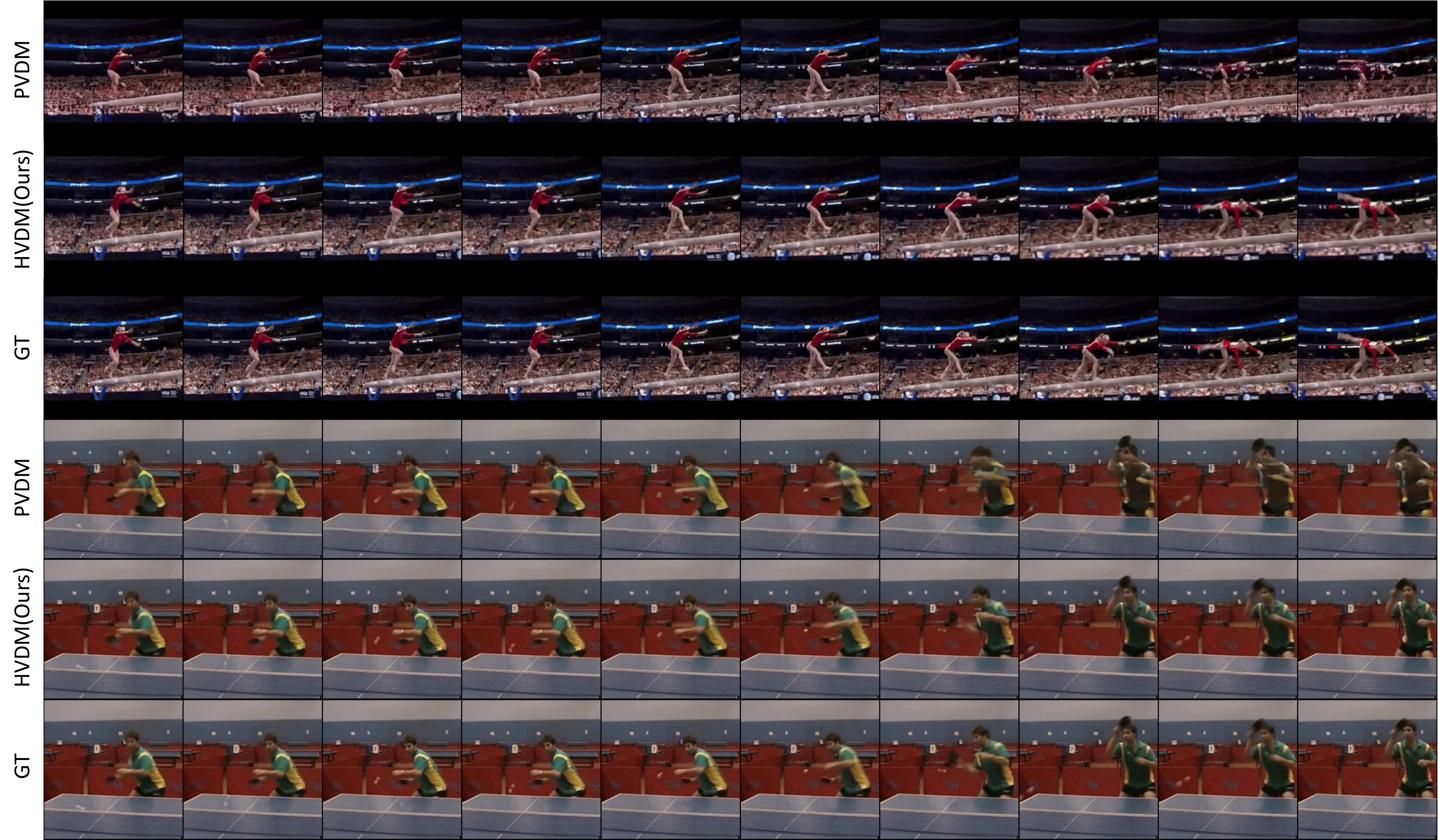}
\captionof{figure}{\textbf{Reconstruction results of our hybrid autoencoder on the UCF-101~\cite{soomro2012ucf101}.}}
\end{figure}

\clearpage
\subsection{Unconditional Video Generation}
\begin{figure*}[h]
\centering
\includegraphics[width=0.85\textwidth]{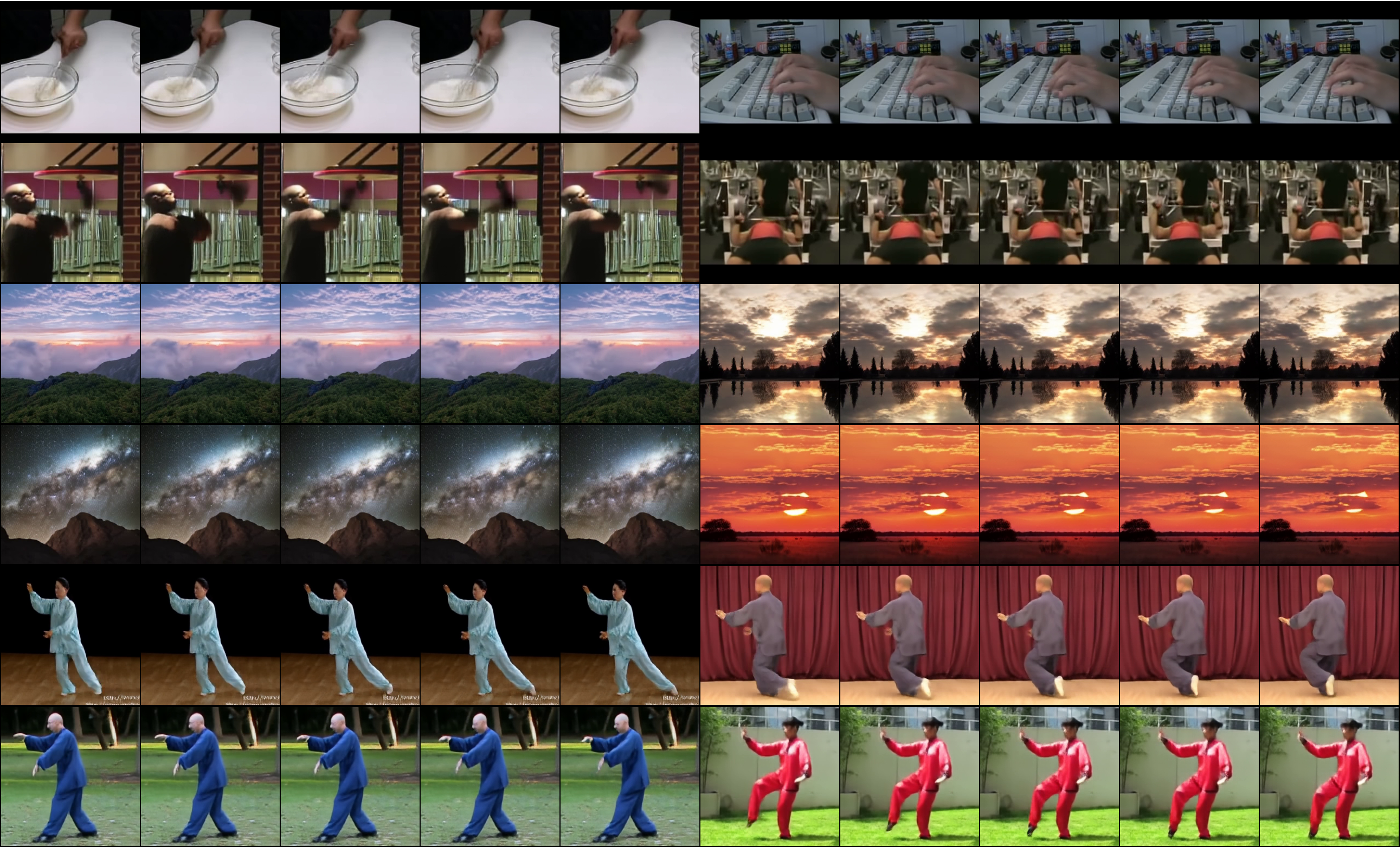}
\end{figure*}
\vspace{-1em}
\captionof{figure}{\textbf{Unconditional generation results of HVDM.}}

\vspace{0.4in}

\subsection{Long Video Generation}
\begin{figure*}[h]
\centering
\includegraphics[width=0.85\textwidth]{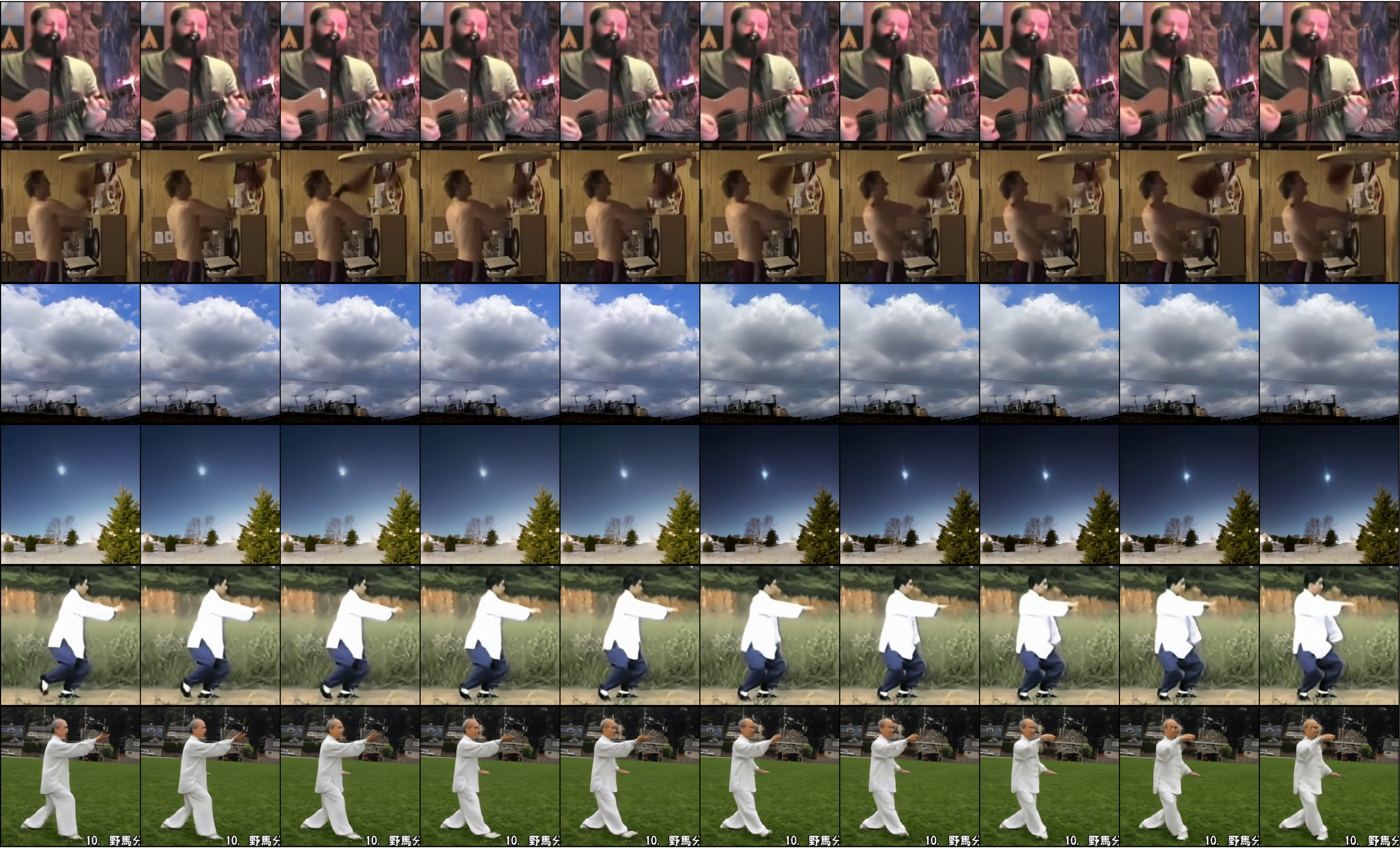}
\end{figure*}
\vspace{-1em}
\captionof{figure}{\textbf{Long video generation results of HVDM.}}
\clearpage

\clearpage
\subsection{Image-To-Video}
\begin{figure*}[h]
\centering
\vspace{-1em}
\includegraphics[width=0.85\textwidth]{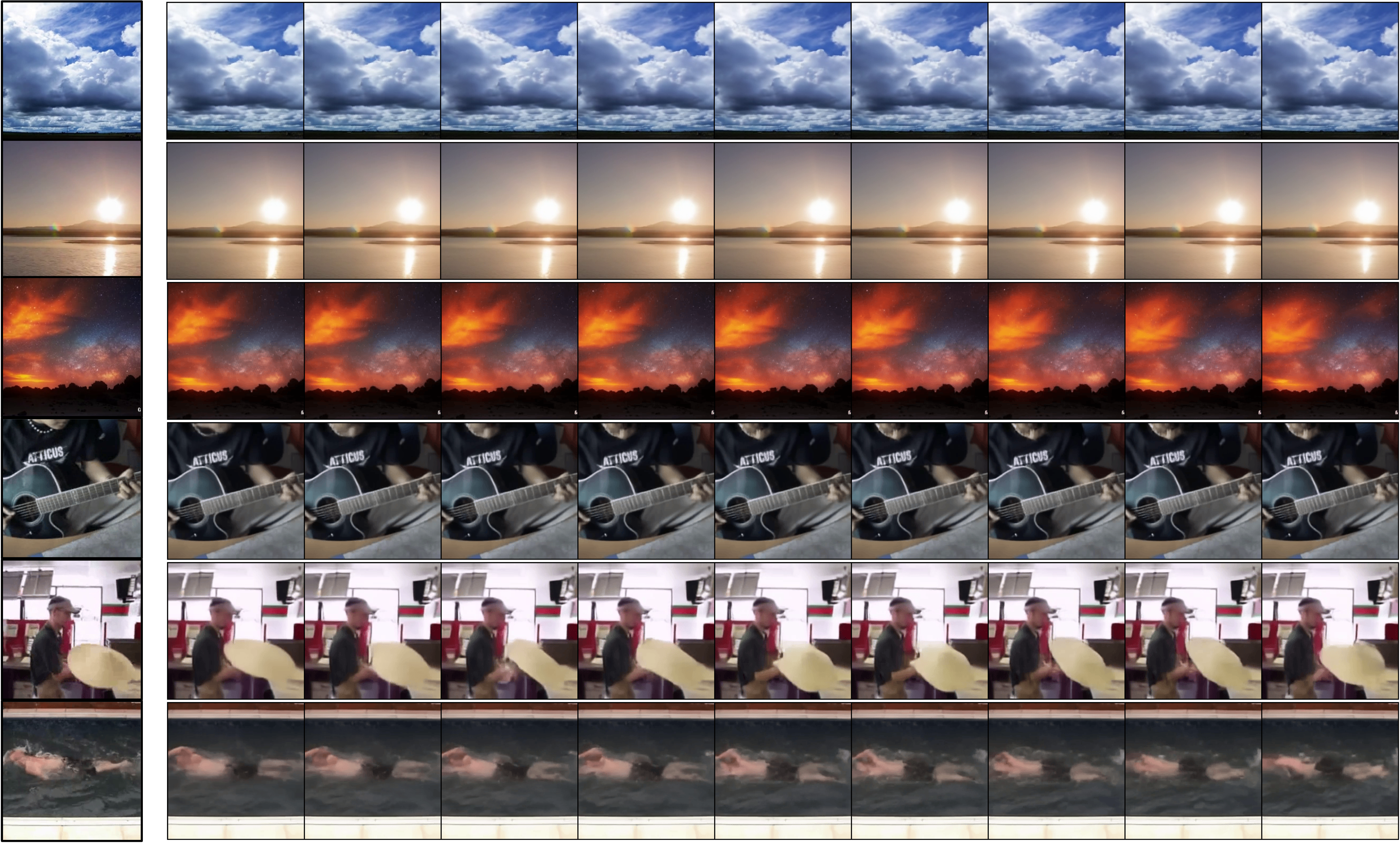}
\end{figure*}
\vspace{-1em}
\captionof{figure}{\textbf{Image-to-video generation results of HVDM.}}

\vspace{0.4in}

\subsection{Video Dynamics Control}
\begin{figure*}[h]
\centering
\includegraphics[width=0.85\textwidth]{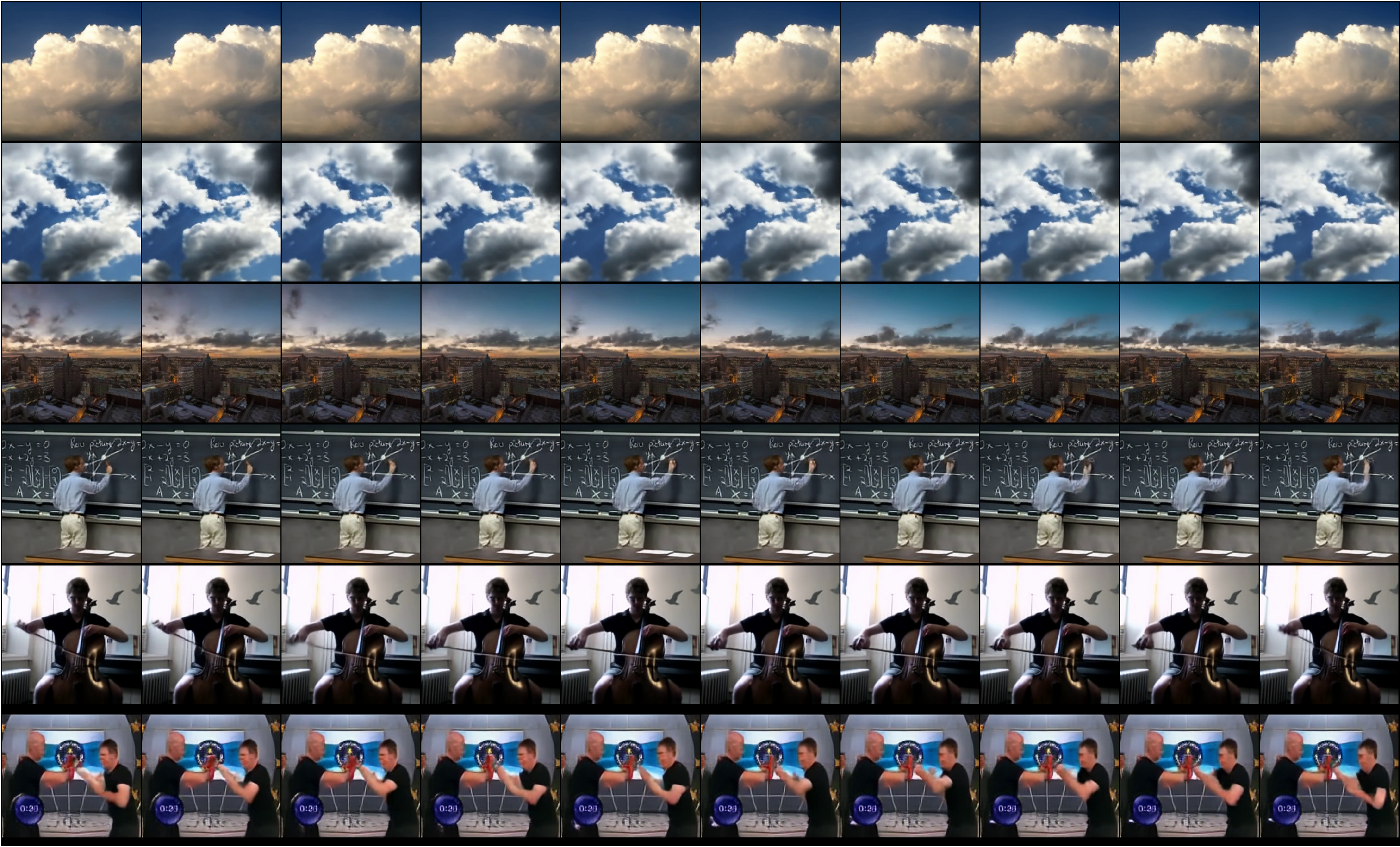}
\end{figure*}
\vspace{-1em}
\captionof{figure}{\textbf{Video dynamics control results of HVDM.}}

\end{document}